\documentclass{article}

\PassOptionsToPackage{numbers,compress}{natbib}

\usepackage[preprint]{nips_2018}

\usepackage[utf8]{inputenc} 
\usepackage[T1]{fontenc}    
\usepackage{hyperref}       
\usepackage{url}            
\usepackage{booktabs}       
\usepackage{amsfonts}       
\usepackage{nicefrac}       
\usepackage{microtype}      

\usepackage{amsmath,amssymb,amsthm,amsbsy}
\usepackage{graphicx}
\usepackage{subcaption}
\usepackage{multirow}
\usepackage{algorithm}
\usepackage{algorithmic}
\usepackage{enumitem}

\theoremstyle{remark}

\newcommand{\round}{\operatorname{round}}

\newcommand{\sign}{\operatorname{sign}}

\title{Compression of Deep Convolutional Neural Networks under Joint Sparsity Constraints}

\author{
  Yoojin Choi\qquad~Mostafa El-Khamy\qquad~Jungwon Lee\\
  SoC R\&D, Samsung Semiconductor Inc.\\
  San Diego, CA 92121, USA\\
  \texttt{\{yoojin.c,mostafa.e,jungwon2.lee\}@samsung.com} \\
}

\begin{document}

\maketitle

\begin{abstract}
We consider the optimization of deep convolutional neural networks (CNNs) such that they provide good performance while having reduced complexity if deployed on either conventional systems utilizing spatial-domain convolution or lower complexity systems designed for Winograd convolution. Furthermore, we explore the universal quantization and compression of these networks. In particular, the proposed framework produces one compressed model whose convolutional filters can be made sparse either in the spatial domain or in the Winograd domain. Hence, one compressed model can be deployed universally on any platform, without need for re-training on the deployed platform, and the sparsity of its convolutional filters can be exploited for further complexity reduction in either domain. To get a better compression ratio, the sparse model is compressed in the spatial domain which has a less number of parameters. From our experiments, we obtain $24.2\times$, $47.7\times$ and $35.4\times$ compressed models for ResNet-18, AlexNet and CT-SRCNN, while their computational cost is also reduced by $4.5\times$, $5.1\times$ and $23.5\times$, respectively.
\end{abstract}

\section{Introduction} \label{sec:intro}

Deep learning with convolutional neural networks (CNNs) has recently achieved performance breakthroughs in many of computer vision applications~\citep{lecun2015deep}. 
However, the large model size and huge computational complexity hinder the deployment of state-of-the-art CNNs on resource-limited platforms such as battery-powered mobile devices. Thus, it is of great interest to compress large-size CNNs into compact forms to lower their storage requirements and to reduce their computational costs~\citep{sze2017efficient,cheng2018model}.

CNN size compression has been actively investigated for memory and storage size reduction. \citet{han2015deep} showed impressive compression results by weight pruning, quantization using $k$-means clustering and Huffman coding. It has been followed by further analysis and mathematical optimization, and more efficient CNN compression schemes have been suggested afterwards, e.g., in \citet{choi2017towards,ullrich2017soft,agustsson2017soft,molchanov2017variational,louizos2017bayesian,choi2018universal,dai2018compressing}. 
CNN computational complexity reduction has also been investigated on the other hand. The major computational cost of CNNs comes from the multiply-accumulate (MAC) operations in their convolutional layers~\citep[Table~II]{sze2017efficient}.
There have been two directions to reduce the complexity of convolutions in CNNs:
\begin{itemize}[noitemsep,topsep=0em,leftmargin=1.6em]
\item First, instead of conventional spatial-domain convolution, it is suggested to use either frequency-domain convolution~\citep{mathieu2013fast,vasilache2014fast} or Winograd convolution~\citep{lavin2016fast}. In particular, for typical small-size filters such as $3\times3$ filters, \citet{lavin2016fast} showed that Winograd convolution is more efficient than both spatial-domain convolution and frequency-domain convolution.
\item Second, weight pruning is another approach to reduce the number of MACs required for convolution by skipping the MACs involving pruned weights (zero weights). The previous work mostly focused on spatial-domain weight pruning, which leads us to exploit sparse spatial-domain convolution of low complexity, e.g., see \citet{han2015learning,lebedev2016fast,wen2016learning,guo2016dynamic,lin2017runtime,park2017faster}. More recently, there have been some attempts to prune Winograd-domain weights and reduce the complexity of Winograd convolution~\citep{li2017enabling,liu2018efficient}.
\end{itemize}


Previous works either focused on spatial-domain weight pruning and compression or focused on Winograd-domain weight pruning and complexity reduction. Compression of Winograd CNNs has never been addressed before, to the best of our knowledge. Other shortcomings of the previous works addressing the complexity reduction of Winograd CNNs are that their final CNNs are no longer backward compatible with spatial-domain convolution due to the non-invertibility of the Winograd transformation, and hence they suffer from accuracy losses if they need to be run on platforms that only support spatial-domain convolution. To our knowledge, this paper is the first to address the universal CNN pruning and compression framework for both Winograd and spatial-domain convolutions.

\begin{figure}[t]
\centering
\includegraphics[width=\textwidth]{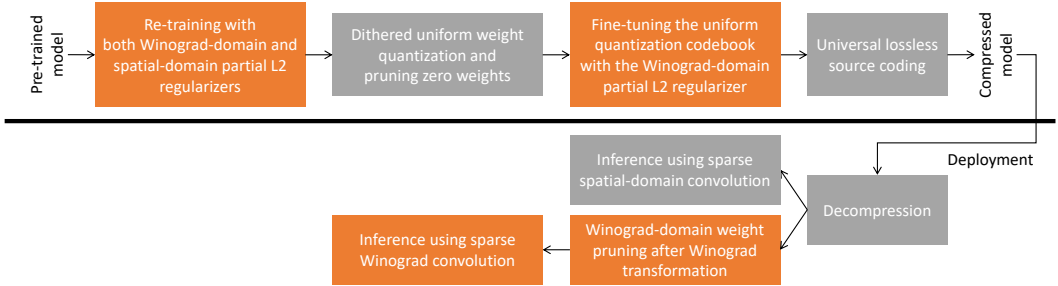}
\caption{Universal CNN weight pruning and compression for supporting sparse Winograd convolution as well as sparse spatial-domain convolution.\label{sec:intro:fig:01}}
\end{figure}

Our proposed solutions are summarized in Figure~\ref{sec:intro:fig:01}. The main novelty of the proposed framework comes from the fact that it optimizes CNNs such their convolutional filters can be pruned either in the Winograd domain or in the spatial domain without accuracy losses and without extra training or fine-tuning in that domain. Our CNNs can be further optimized for and compressed by universal quantization and universal source coding such that their decompressed convolutional filters still have sparsity in both Winograd and spatial domains. Hence, one universally compressed model can be deployed on any platform whether it uses spatial-domain convolution or Winograd convolution, and the sparsity of its convolutional filters can be utilized for complexity reduction in either domain, with no need for further training. Since many low-power platforms, such as mobile phones, are expected to only support the inference of CNNs, and not their training, our approach is more desirable for wide-scale deployment of pre-trained models without worrying about underlying inference engines.

\section{Preliminary}

\subsection{Winograd convolution} \label{sec:winograd}

We first review the Winograd convolution algorithm~\citep{winograd1980arithmetic} in this subsection. It is well known that spatial-domain convolution is equivalent to element-wise product in the frequency domain or in the Winograd domain (e.g., see \citet[Section~5]{blahut2010fast}). In particular, the Winograd convolution algorithm is designed to compute a convolution with the minimum multiplications possible~\citep[Section~8.4]{selesnick1998fast}.

For the sake of illustration, consider that we are given a two-dimensional (2-D) input of size $H\times W$ and a 2-D filter of size $r\times r$ for convolution. We first prepare a set of patches of size $n\times n$ extracted from the input with stride of $n-r+1\times n-r+1$ for $n\geq r$. Each of the $n\times n$ patches is convolved with the $r\times r$ filter by the Winograd convolution algorithm and produces an output patch of size $n-r+1\times n-r+1$. Finally, the output patches are combined into one output image. 

Let $x$ and $y$ be one of the $n\times n$ input patches and its corresponding output patch, respectively, and let $w$ be the $r\times r$ filter. In Winograd convolution, the input and the filter are transformed into the Winograd domain by $X=FxF^T$ and $W=GwG^T$ using the Winograd transformation matrices~$F$ and $G$, respectively, where the superscript~$T$ denotes the matrix transpose. In the Winograd domain, both $X$ and $W$ are of size $n\times n$, and element-wise product of them follows. Then, the output is transformed back to the spatial domain using matrix~$S$ by
\begin{equation} \label{sec:winograd:eq:01}
y=S^T(W\odot X)S,
\end{equation}
where $\odot$ is the element-wise product of two matrices. The transformation matrices $F$, $G$, and $S$ are $(r,n)$-specific and can be obtained from the Chinese remainder theorem (e.g., see \citet[Section~5.3]{blahut2010fast}). In case of $C$ input channels, the inverse transformation in \eqref{sec:winograd:eq:01} can be deployed once after summation over all channels of the element-wise product outputs in the Winograd domain (see \citet[Section~4]{lavin2016fast}), i.e.,
\[
y=S^T\left[\sum_{c=1}^C(W_c\odot X_c)\right]S,
\]
where $W_c$ and $X_c$ are the Winograd-transformed filter and input of channel~$c$, respectively (see \citet{lavin2016fast}).



\subsection{Sparse Winograd convolution} \label{sec:sparse}

Similar to spatial-domain weight pruning for sparse spatial-domain convolution of low complexity, it is considered to skip some of the computations in the Winograd domain by pruning (i.e., setting to zero) some of the Winograd-transformed filter weights (elements of $W$ in \eqref{sec:winograd:eq:01}) for sparse Winograd convolution. The most related work to this end can be found in \citet{li2017enabling,liu2018efficient}.

Pruning spatial-domain weights does not yield sparse Winograd-domain filters in general since the sparsity is not maintained after transformation. Thus, \citet{li2017enabling} introduced new Winograd layers, which are similar to convolutional layers but their learnable parameters are defined in the Winograd domain, and not in the spatial domain. In their framework, Winograd-domain weights are directly learned in training where the loss and gradients are computed with Winograd layers. For Winograd-domain weight pruning, some insignificant Winograd-domain weights are nullified in every training iteration based on their magnitude and gradient values. In \citet{liu2018efficient}, the complexity of Winograd layers is further reduced by putting rectified linear units (ReLUs) in the Winograd domain and skipping MACs not only for zero weights, but also for zero activations in the Winograd domain.


However, if we learn Winograd-domain weights directly using Winograd layers, the trained model has to use Winograd layers in inference as well. We cannot transform the learned Winograd-domain weights back to the spatial domain without considerable accuracy loss, since the inverse transformation from the Winograd domain to the spatial domain is over-determined. Hence, the model is not deployable on the platforms that only support classical spatial-domain convolution. Moreover, storing Winograd-domain weights is inefficient, since the number of weights is larger in the Winograd domain. Thus, we suggest that it is better to compress weights in the spatial domain even if the target computational platform only deploys Winograd convolution.


\subsection{Universal compression}

A universal CNN compression framework was proposed in \cite{choi2018universal}, where CNNs are optimized for and compressed by universal quantization and universal entropy source coding with schemes such as the variants of Lempel--Ziv--Welch~\citep{ziv1977universal,ziv1978compression,welch1984technique} and the Burrows--Wheeler transform~\citep{effros2002universal}. Of particular interest for universal quantization is randomized uniform quantization, where uniform random dithering makes the distortion independent of the source, and the gap of its rate from the rate-distortion bound at any distortion level is provably no more than $0.754$ bits per sample for any source~\citep{zamir1992universal}. Universal CNN compression has practical advantages as it is easily applicable to any CNN model at any desired compression rate, without the extra burden required by previous approaches to compute or estimate the statistics of the CNN weights, and is guaranteed to achieve near-optimal performance.

\section{Training with joint sparsity constraints}

In this section, we present our CNN training method with regularization under joint spatial-Winograd sparsity constraints, to enable efficient deployment of pre-trained CNNs in either domain, without additional training for deployment.

\subsection{CNN model}

We consider a typical CNN model consisting of $L$ convolutional layers. The input of layer~$l$ has $C_l$ channels of size $H_l\times W_l$ and the output has $D_l$ channels of size $H_l-r_l+1\times W_l-r_l+1$, where the input is convolved with $D_l$ filters of size $r_l\times r_l\times C_l$. For $1\leq l\leq L$, $1\leq i\leq C_l$ and $1\leq j\leq D_l$, 
let $w_l(i,j)$ be the 2-D convolutional filter for input channel~$i$ and output channel~$j$ of layer~$l$. 

\subsection{Regularization for jointly sparse convolutional filters} \label{sec:reg}

In this subsection, we introduce our Winograd-domain and spatial-domain partial L2 regularizers to attain convolutional filters that are sparse in both the Winograd domain and the spatial domain. We choose L2 regularizers to promote sparsity, although other regularizers such as L1 regularizers can be used instead (see Section~\ref{sec:discussion} for more discussion). For notational simplicity, let $\mathbf{w}$ be the set of all convolutional filters of $L$ layers, which are learnable, i.e.,
\[
\mathbf{w}\equiv\{w_l(i,j),1\leq l\leq L,1\leq i\leq C_l,1\leq j\leq D_l\}.
\]
Moreover, given any matrix~$A$, we define $1_{|A|\leq\theta}$ as the matrix that has the same size as $A$ while its element is one if the corresponding element~$a$ in $A$ satisfies $|a|\leq\theta$ and is zero otherwise.

\textbf{Winograd-domain partial L2 regularization}: To optimize CNNs under a Winograd-domain sparsity constraint, we introduce the Winograd-domain partial L2 regularizer given by
\begin{equation} \label{sec:reg:eq:01}
R_{\text{WD}}(\mathbf{w};s_{\text{WD}})
=\frac{1}{N_{\text{WD}}}\sum_{l=1}^L\sum_{i=1}^{C_l}\sum_{j=1}^{D_l}\|(G_lw_l(i,j)G_l^T)\odot1_{|G_lw_l(i,j)G_l^T|\leq\theta_{\text{WD}}(s_{\text{WD}})}\|^2,
\end{equation}
where $\|\cdot\|$ denotes the L2 norm and $G_l$ is the Winograd transformation matrix determined by the filter size and the input patch size of layer~$l$ for Winograd convolution (see Section~\ref{sec:winograd}); $N_{\text{WD}}$ is the total number of Winograd-domain weights of all $L$ layers. Although the constraints are on the Winograd-domain weights, they translate as the constraints on the corresponding spatial-domain weights, and the optimization is done in the spatial domain; this facilitates the optimization for additional sparsity constraints in the spatial domain as will be clarified below.

Observe that the L2 regularization in \eqref{sec:reg:eq:01} is applied only to a part of Winograd-domain weights if their magnitude values are not greater than the threshold value~$\theta_{\text{WD}}(s_{\text{WD}})$. Due to the partial L2 regularization, spatial-domain weights are updated towards the direction to yield diminishing Winograd-domain weights in part after training and being transformed into the Winograd domain. 
Given a desired sparsity level~$s_{\text{WD}}$ (\%) in the Winograd domain, we set the threshold value~$\theta_{\text{WD}}(s_{\text{WD}})$ to be the $s_{\text{WD}}$-th percentile of Winograd-domain weight magnitude values. The threshold is updated at every training iteration as weights are updated. Note that the threshold decreases as training goes on since the regularized Winograd-domain weights gradually converge to small values in magnitude (see Section~\ref{sec:train}). After finishing the regularized training, we finally have a set of Winograd-domain weights clustered very near zero, which can be pruned (i.e., set to zero) at minimal accuracy loss.

\textbf{Spatial-domain partial L2 regularization}: To optimize CNNs while having sparsity in the spatial domain, we regularize the cost function by the partial sum of L2 norms of spatial-domain weights, determined by $\theta_{\text{SD}}(s_{\text{SD}})$ given a target sparsity level~$s_{\text{SD}}$ (\%), similar to \eqref{sec:reg:eq:01}, as below:
\begin{equation} \label{sec:reg:eq:02}
R_{\text{SD}}(\mathbf{w};s_{\text{SD}})
=\frac{1}{N_{\text{SD}}}\sum_{l=1}^L\sum_{i=1}^{C_l}\sum_{j=1}^{D_l}\|w_l(i,j)\odot1_{|w_l(i,j)|\leq\theta_{\text{SD}}(s_{\text{SD}})}\|^2,
\end{equation}
where $N_{\text{SD}}$ is the total number of spatial-domain weights of all $L$ layers.

\subsection{Regularized training with learnable regularization coefficients} \label{sec:train}

Combining the regularizers in \eqref{sec:reg:eq:01} and \eqref{sec:reg:eq:02}, the cost function~$C$ to minimize in training is given by
\begin{equation} \label{sec:reg:eq:03}
C(\mathcal{X};\mathbf{w})=E(\mathcal{X};\mathbf{w})+\lambda_{\text{WD}} R_{\text{WD}}(\mathbf{w};s_{\text{WD}})+\lambda_{\text{SD}} R_{\text{SD}}(\mathbf{w};s_{\text{SD}}),
\end{equation}
for $\lambda_{\text{WD}}>0$ and $\lambda_{\text{SD}}>0$, where $\mathcal{X}$ is a training dataset and the $E$ is the network loss function such as the cross-entropy loss for classification or the mean-squared-error loss for regression. We emphasize that training is performed in the spatial domain with conventional spatial-domain convolution and we update spatial-domain filters in $\mathbf{w}$, while the regularizers steer the filters to have a desired percentage of weights with small or near-zero values either in the spatial domain or in the Winograd domain when transformed, which are safe to prune at little accuracy loss.

In \eqref{sec:reg:eq:03}, we introduce two regularization coefficients~$\lambda_{\text{SD}}$ and $\lambda_{\text{WD}}$. Conventionally, we use a fixed value for a regularization coefficient. However, we observe that using fixed regularization coefficients for the whole training is not efficient to find good sparse models. For small fixed coefficients, regularization is weak and we cannot reach the desired sparsity after training.
For large fixed coefficients, on the other hand, we can achieve the desired sparsity, but it likely comes with considerable performance loss due to strong regularization.

\textbf{Learnable regularization coefficient}: To overcome the problems with fixed regularization coefficients, we propose novel \emph{learnable regularization coefficients}, i.e., we let regularization coefficients be learnable parameters. Starting from a small initial coefficient value, we learn an accurate model with little regularization. As training goes on, we make the regularization coefficients increase gradually so that the performance does not degrade much but we finally have sparse convolutional filters at the end of training in both Winograd and spatial domains. Towards this end, we first replace $\lambda_{\text{WD}}$ and $\lambda_{\text{SD}}$ with $e^{\zeta_{\text{WD}}}$ and $e^{\zeta_{\text{SD}}}$, respectively, and learn $\zeta_{\text{WD}}$ and $\zeta_{\text{SD}}$ instead, for the sake of guaranteeing that the regularization coefficients always positive in training. Moreover, we include an additional regularization term, e.g., $-\alpha(\zeta_{WD}+\zeta_{SD})$ for $\alpha>0$, to penalize small regularization coefficients and encourage them to increase in training. The cost function in \eqref{sec:reg:eq:03} is then altered into
\begin{equation} \label{sec:reg:eq:04}
C(\mathcal{X};\mathbf{w},\zeta_{\text{WD}},\zeta_{\text{SD}})
=E(\mathcal{X};\mathbf{w})
+e^{\zeta_{\text{WD}}}R_{\text{WD}}(\mathbf{w};s_{\text{WD}})
+e^{\zeta_{\text{SD}}}R_{\text{SD}}(\mathbf{w};s_{\text{SD}})
-\alpha(\zeta_{\text{WD}}+\zeta_{\text{SD}}).
\end{equation}
Observe that we introduced a new hyper-parameter~$\alpha$, while making regularization coefficients learnable. The trade-off between the loss and the regularization is now controlled by the new hyper-parameter~$\alpha$ instead of regularization coefficients, which is beneficial since $\alpha$ is not directly related to either of the loss or the regularization, and we can induce smooth transition to a sparse model.

L2 regularization for parameters corresponds to assuming a zero-mean Gaussian prior on the parameters (e.g., see \cite[Section~5.5]{bishop2006pattern}). The Winograd-domain partial L2 regularization can be interpreted as if we assume a zero-mean Gaussian prior for partial Winograd-domain weights within the threshold value and use the negative log-likelihood of the Gaussian prior as a regularization term. The regularization coefficient~$e^{\zeta_{\text{WD}}}$ in \eqref{sec:reg:eq:04} can be related to the variance of the Gaussian prior, i.e., the reciprocal of the variance of the Gaussian prior corresponds to the regularization coefficient~$e^{\zeta_{\text{WD}}}$. In this Bayesian model, we can even consider the variance of the Gaussian prior as a random variable and find the optimal variance by learning, which leads us to the learnable regularization coefficient idea with the penalty term in \eqref{sec:reg:eq:04}. 
A similar interpretation applies to the spatial-domain partial L2 regularization. Training with Gaussian priors has been considered in other contexts, e.g., Gaussian mixture is used for weight quantization in \citet{ullrich2017soft}.

\textbf{Gradient descent}: From \eqref{sec:reg:eq:04}, we have
\begin{equation} \label{sec:reg:eq:05}
\nabla_{w_l(i,j)}C
=\nabla_{w_l(i,j)}E
+e^{\zeta_{\text{WD}}}\nabla_{w_l(i,j)}R_{\text{WD}}
+e^{\zeta_{\text{SD}}}\nabla_{w_l(i,j)}R_{\text{SD}},
\end{equation}
where $\nabla_{w_l(i,j)}E$ is provided from the CNN back-propagation algorithm. It can be shown that 
\begin{gather}
\nabla_{w_l(i,j)}R_{\text{WD}}
=2G_l^T((G_lw_l(i,j)G_l^T)\odot1_{|G_lw_l(i,j)G_l^T|\leq\theta_{\text{WD}}(s_{\text{WD}})})G_l,\label{sec:reg:eq:06}\\
\nabla_{w_l(i,j)}R_{\text{SD}}
=2w_l(i,j)\odot1_{|w_l(i,j)|\leq\theta_{\text{SD}}(s_{\text{SD}})}.\label{sec:reg:eq:07}
\end{gather}
The detailed proof of \eqref{sec:reg:eq:06} is can be found in Appendix, while \eqref{sec:reg:eq:07} is straightforward to show. We note that the indicator functions in \eqref{sec:reg:eq:01} and \eqref{sec:reg:eq:02} are non-differentiable, which is however not a problem when computing the derivatives in practice for stochastic gradient descent. 
Combining \eqref{sec:reg:eq:05}--\eqref{sec:reg:eq:07}, we can perform gradient descent for weights in $w_l(i,j)$. We update $\zeta_{\text{WD}}$ and $\zeta_{\text{SD}}$ by gradient decent using
\[
\nabla_{\zeta_{\text{WD}}}C=e^{\zeta_{\text{WD}}} R_{\text{WD}}-\alpha
\ \ \
\text{and}
\ \ \
\nabla_{\zeta_{\text{SD}}}C=e^{\zeta_{\text{SD}}} R_{\text{SD}}-\alpha,
\]
respectively. Observe that $\zeta_{\text{WD}}$ tends to $\log\alpha-\log{R_{\text{WD}}}$. This implies that as the regularizer~$R_{\text{WD}}$ decreases, the regularization coefficient~$e^{\zeta_{\text{WD}}}$ gets larger. A larger regularization coefficient further encourages spatial-domain weights to move towards the direction where regularized Winograd-domain weights converge zero in the following update. In this way, we gradually sparsify Winograd-domain filters. Similarly, spatial-domain filters are sparsified owing to increasing~$\zeta_{\text{SD}}$ and decreasing~$R_{\text{SD}}$.

\setlength{\tabcolsep}{0.1em}
\begin{figure}[t]
\centering
{\scriptsize
\begin{tabular}{ccccc}
 & (a) Iterations=0 & (b) Iterations=100k & (c) Iterations=120k & (d) Iterations=200k \\
\raisebox{1.5em}{\rotatebox{90}{\shortstack[c]{Winograd domain\\~\\~}}} &
\includegraphics[width=0.23\textwidth]{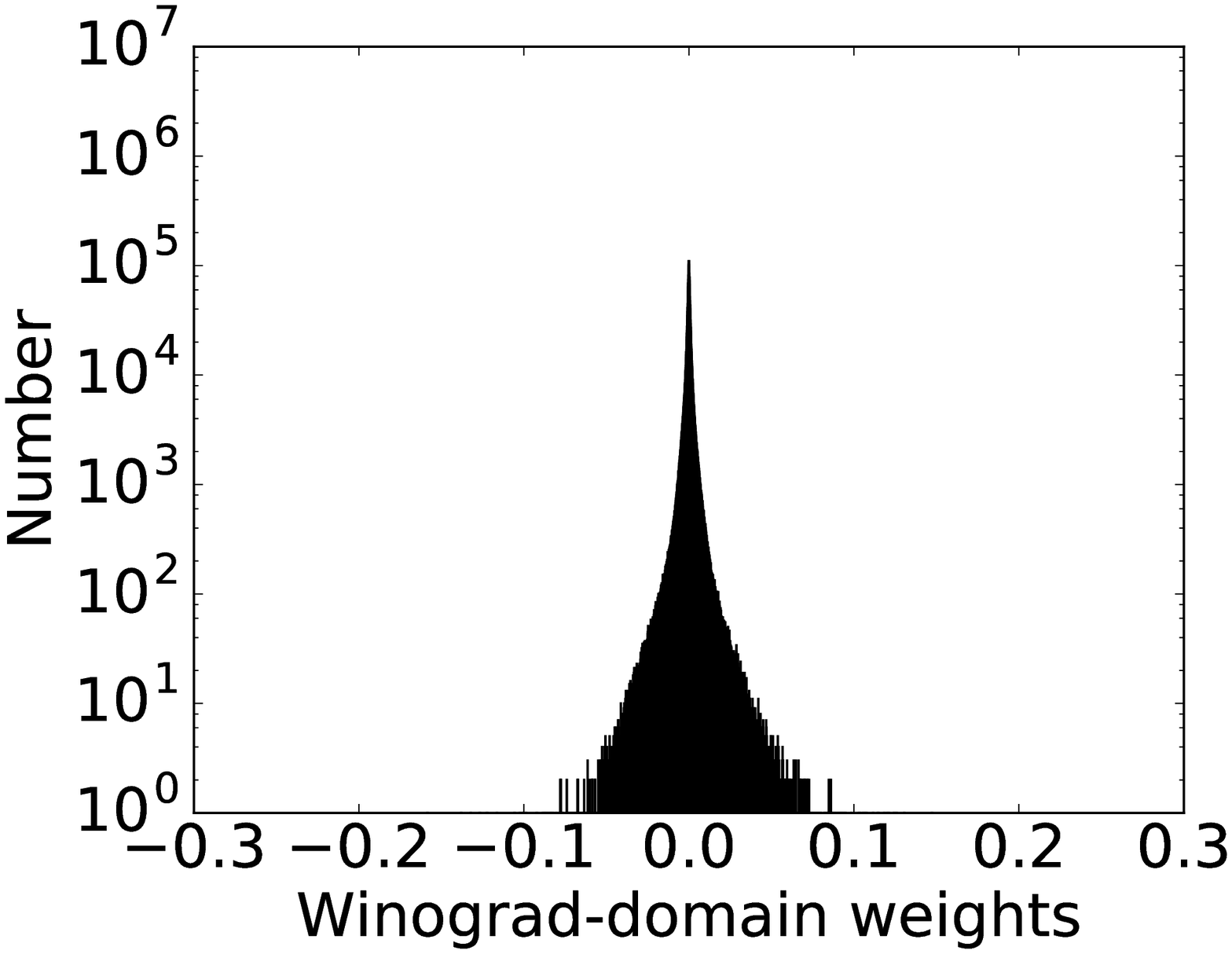} &
\includegraphics[width=0.23\textwidth]{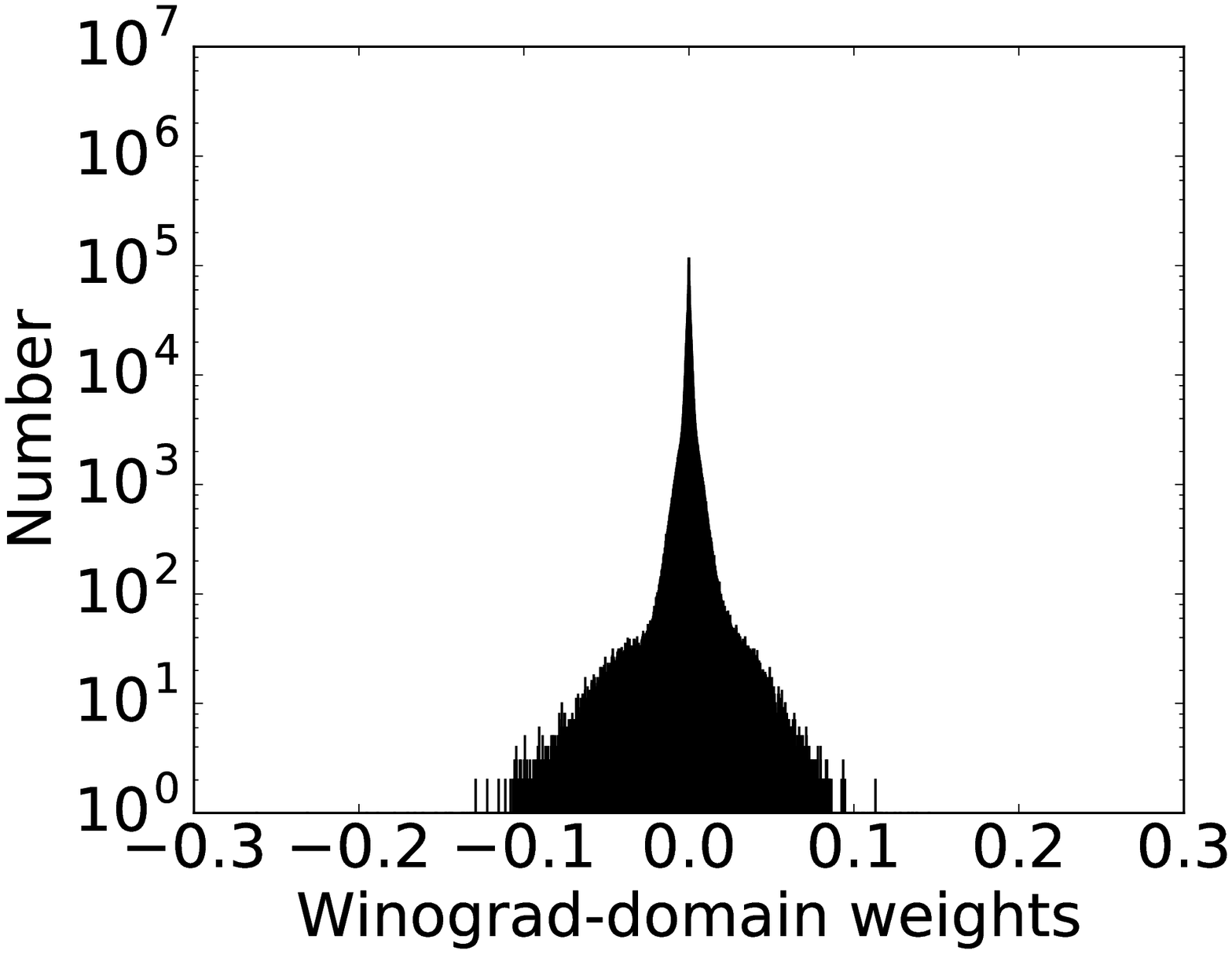} &
\includegraphics[width=0.23\textwidth]{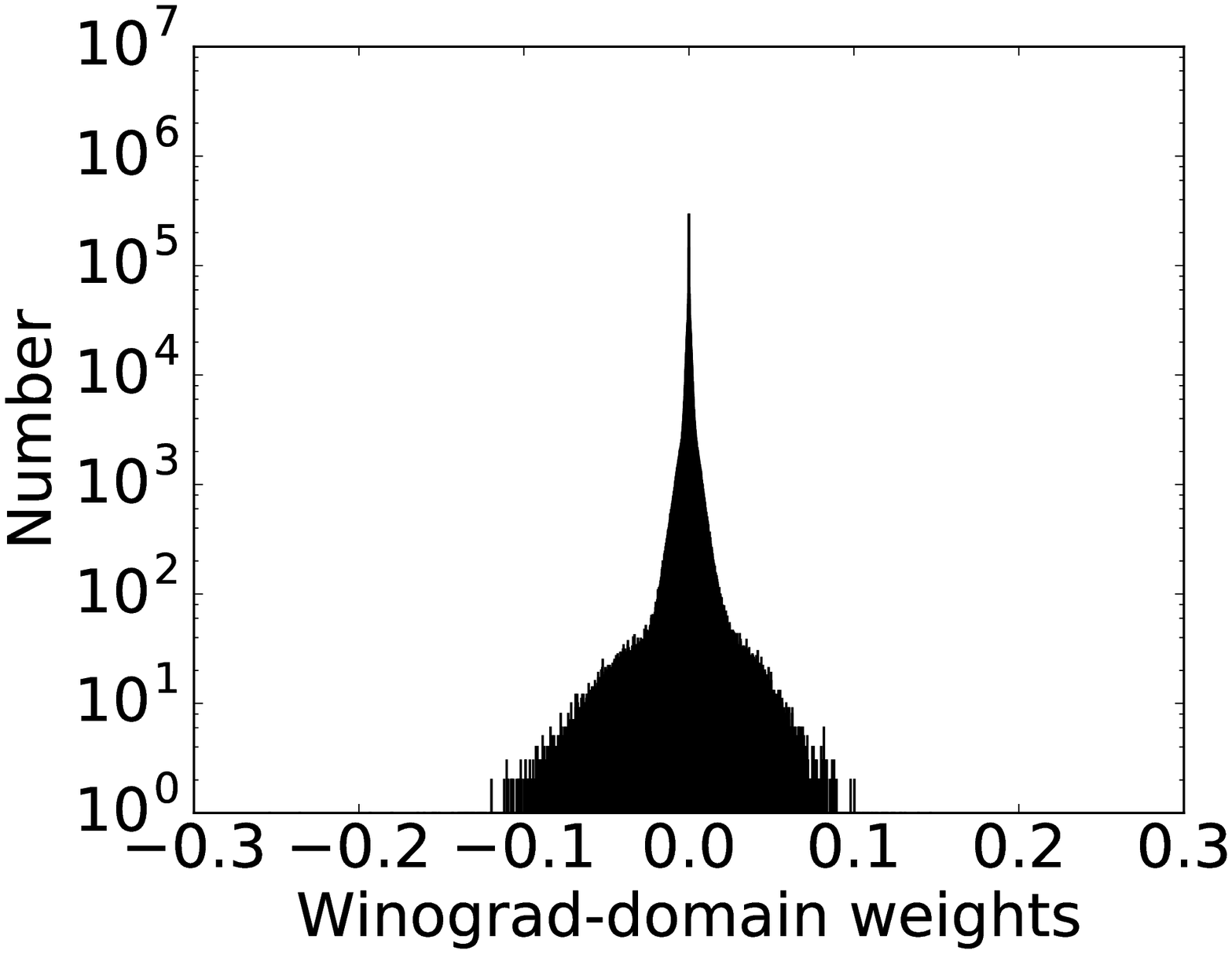} &
\includegraphics[width=0.23\textwidth]{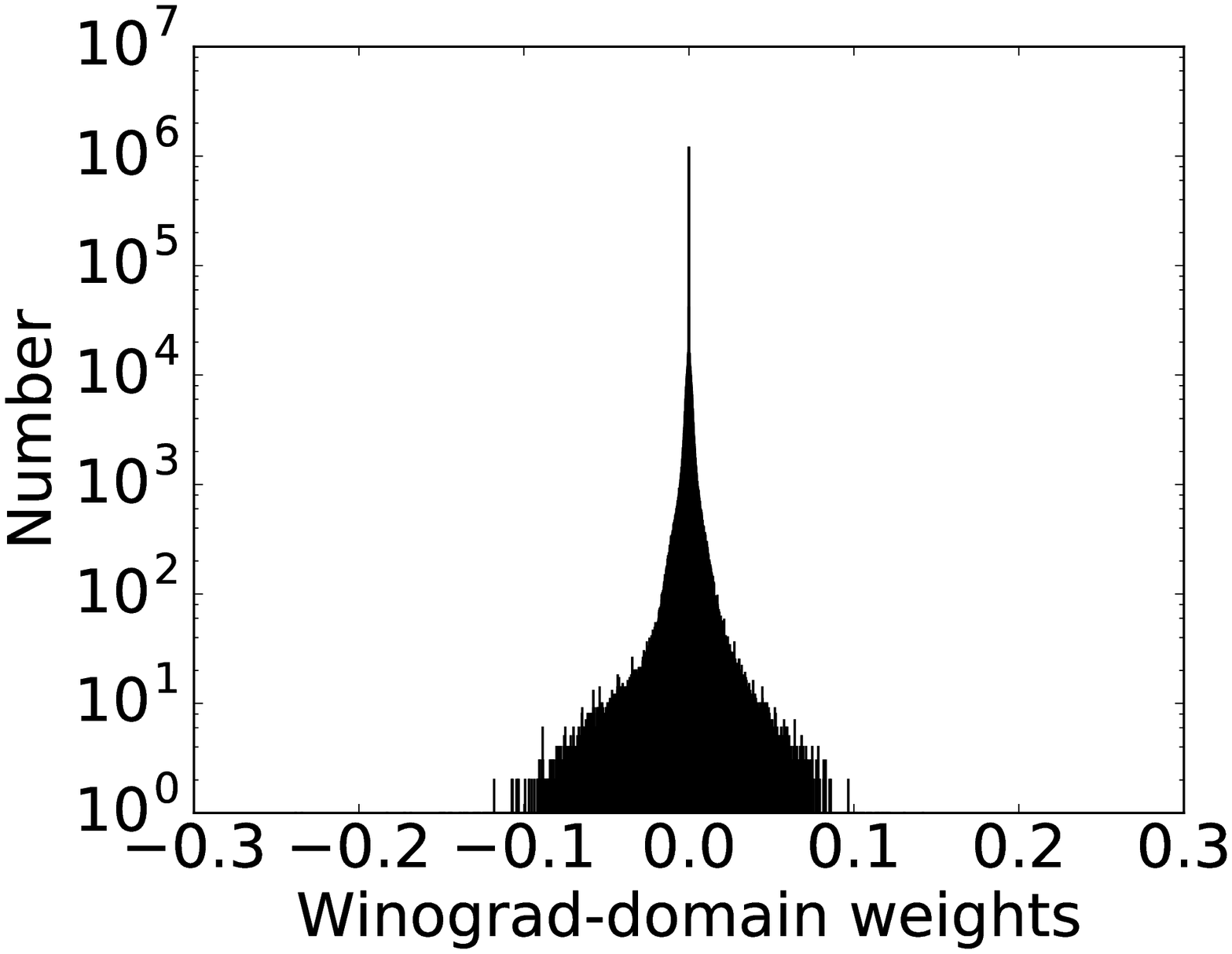} \\
\raisebox{1.5em}{\rotatebox{90}{\shortstack[c]{Spatial domain\\~\\~}}} &
\includegraphics[width=0.23\textwidth]{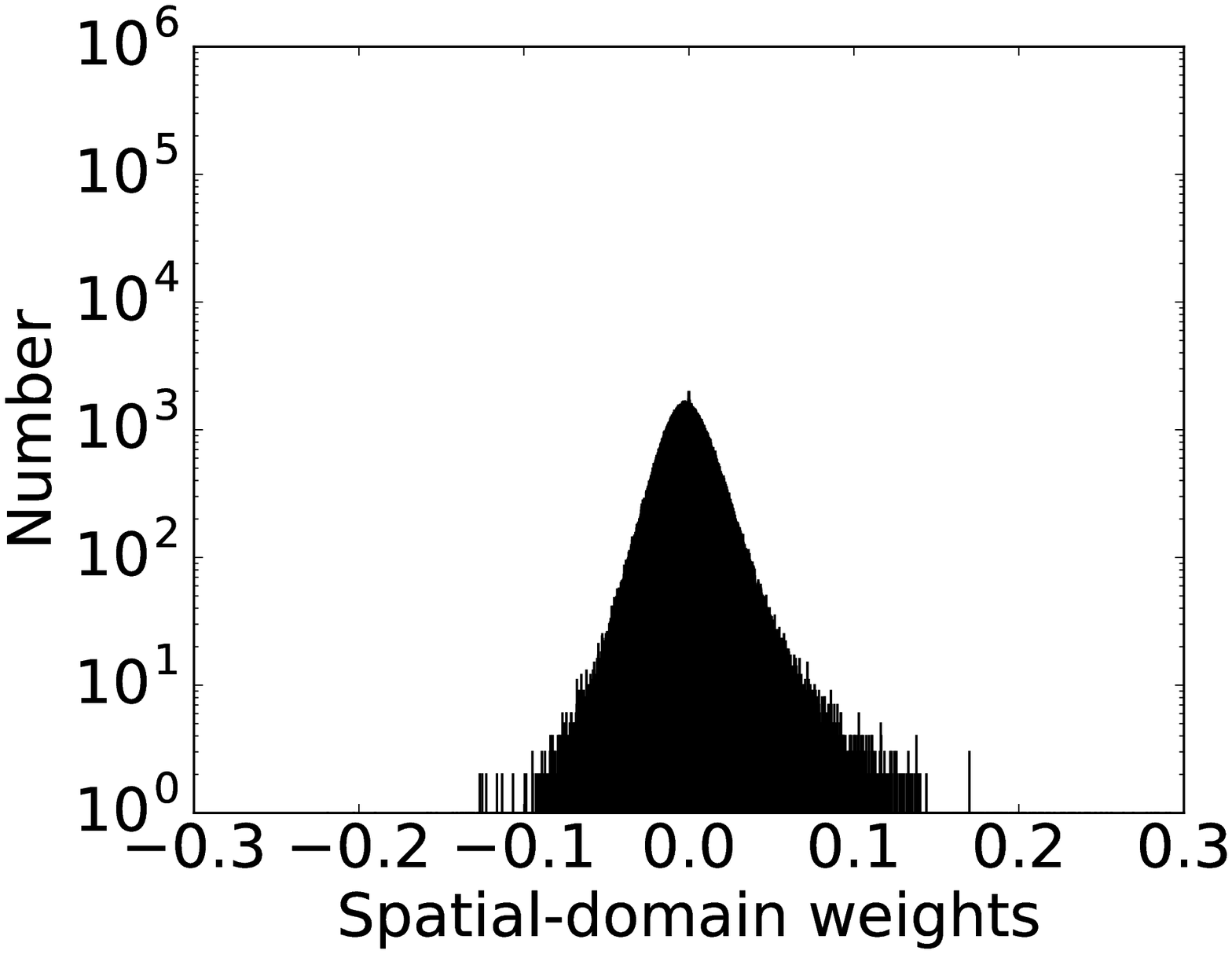} &
\includegraphics[width=0.23\textwidth]{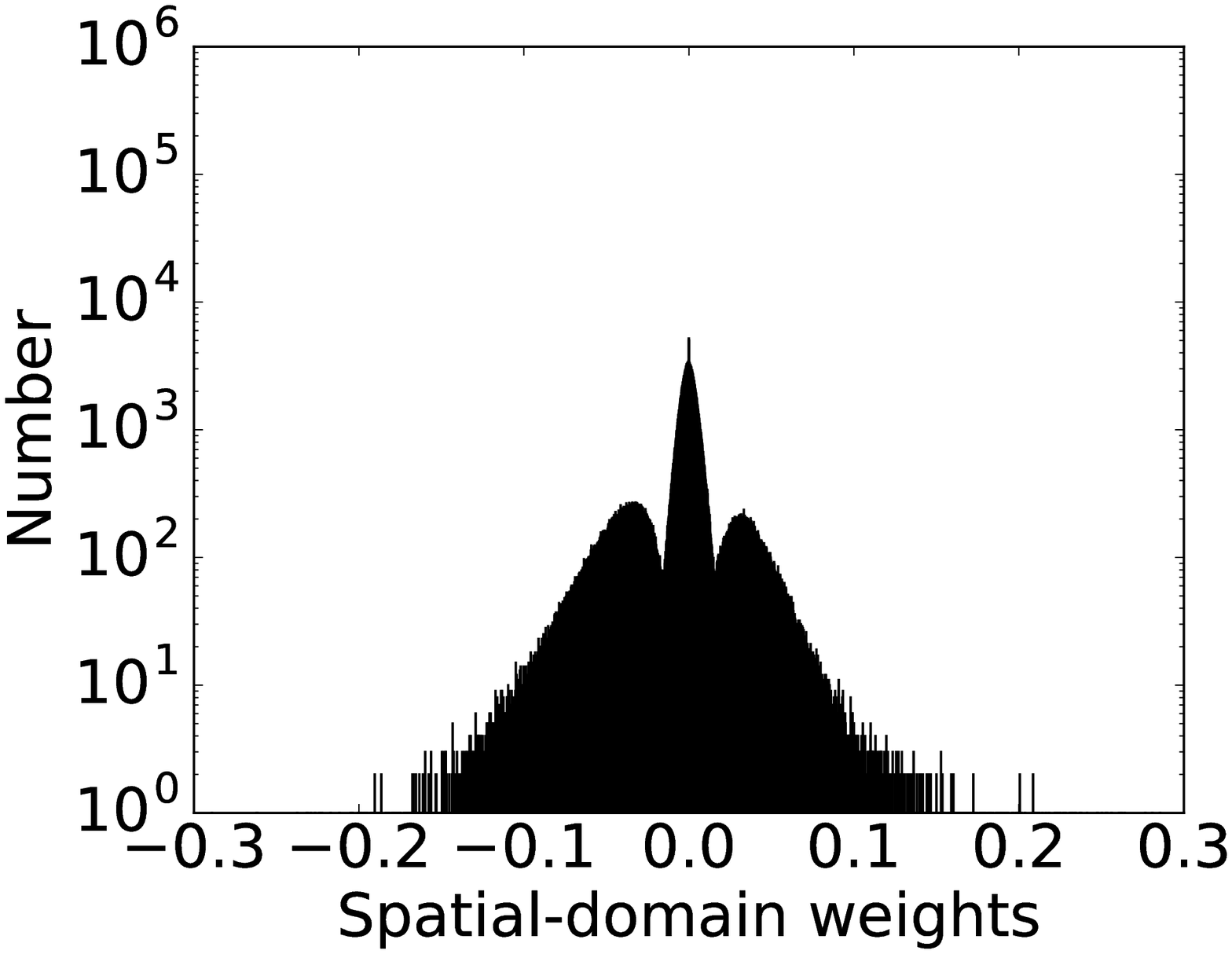} &
\includegraphics[width=0.23\textwidth]{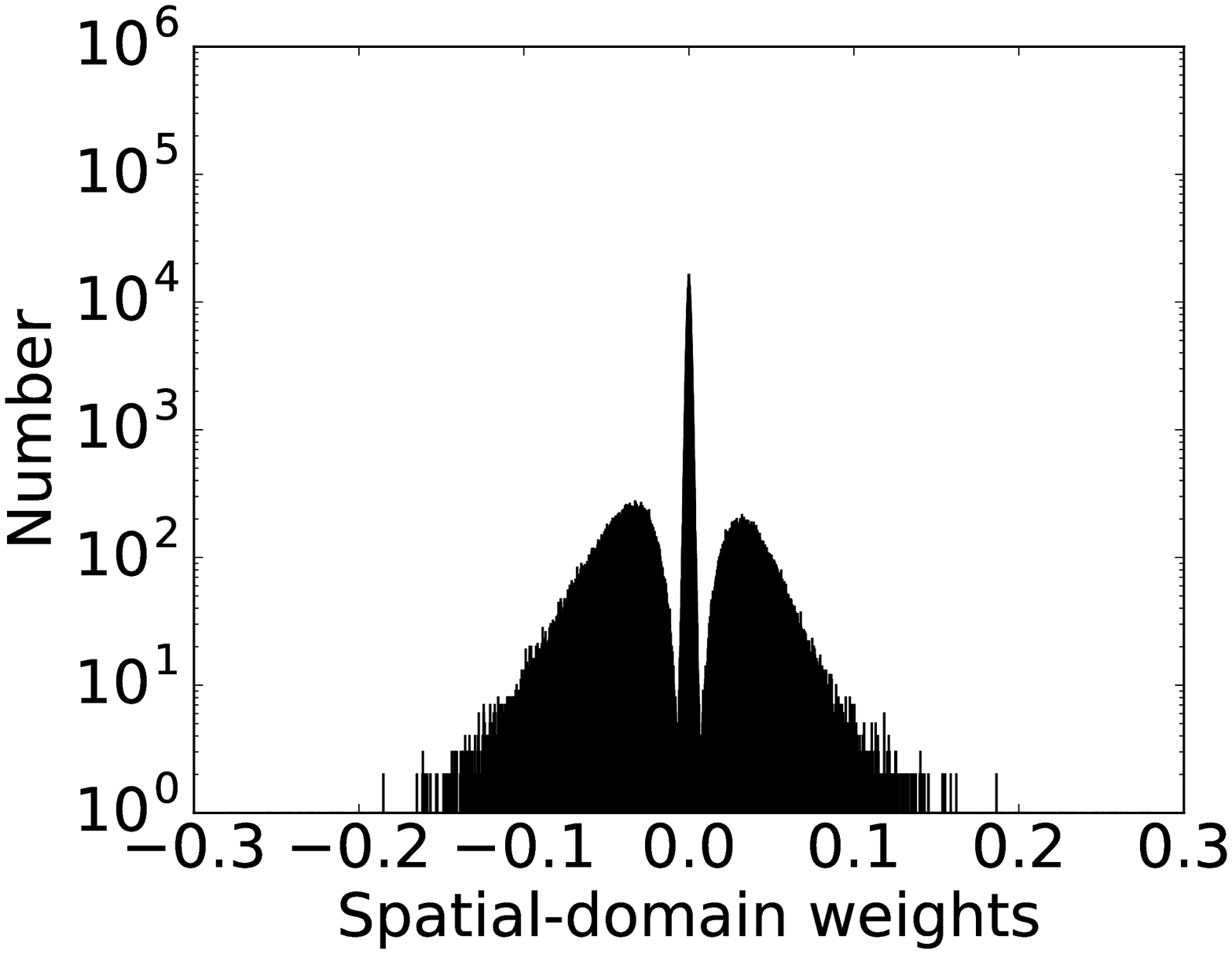} &
\includegraphics[width=0.23\textwidth]{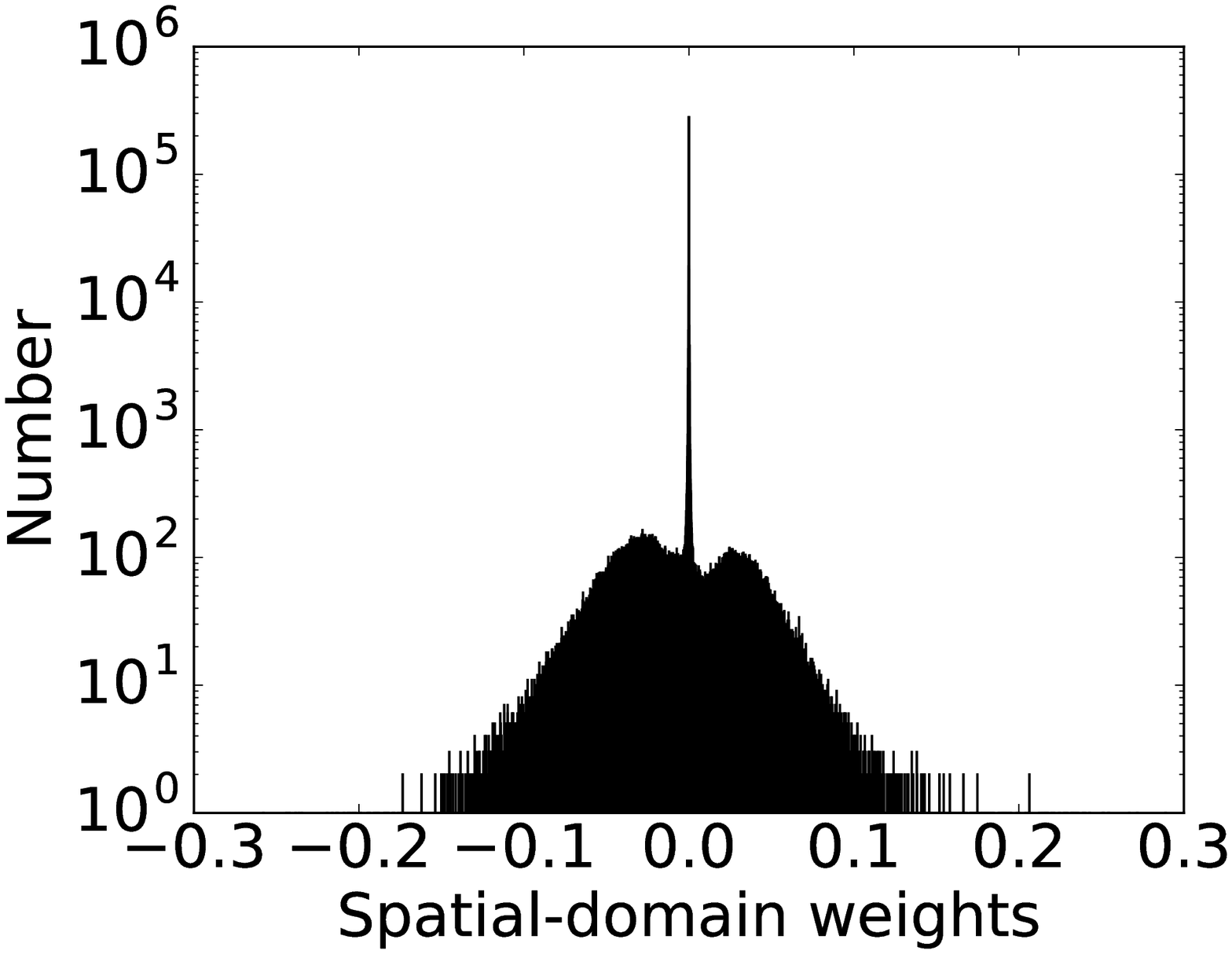} \\
\end{tabular}
}
\caption{Weight histogram snapshots of the AlexNet second convolutional layer.\label{sec:train:fig:01}}
\end{figure}

\textbf{Evolution of weight histogram}: In Figure~\ref{sec:train:fig:01}, we present how the weight histogram (distribution) of the AlexNet second convolutional layer evolves in the Winograd domain and in the spatial domain as training goes on due to the proposed partial L2 regularizers with the learnable regularization coefficients. 
Observe that a part of the weights converges to zero in both domains. Finally, we have a peak at zero, which can be pruned at little accuracy loss, in each domain.

\setlength{\tabcolsep}{0em}
\begin{figure}[t]
\centering
{\scriptsize
\begin{tabular}{llll}
\raisebox{0.2em}{\rotatebox{90}{\shortstack[c]{Winograd\\domain\\~\\~}}} &
\includegraphics[height=0.09\textwidth]{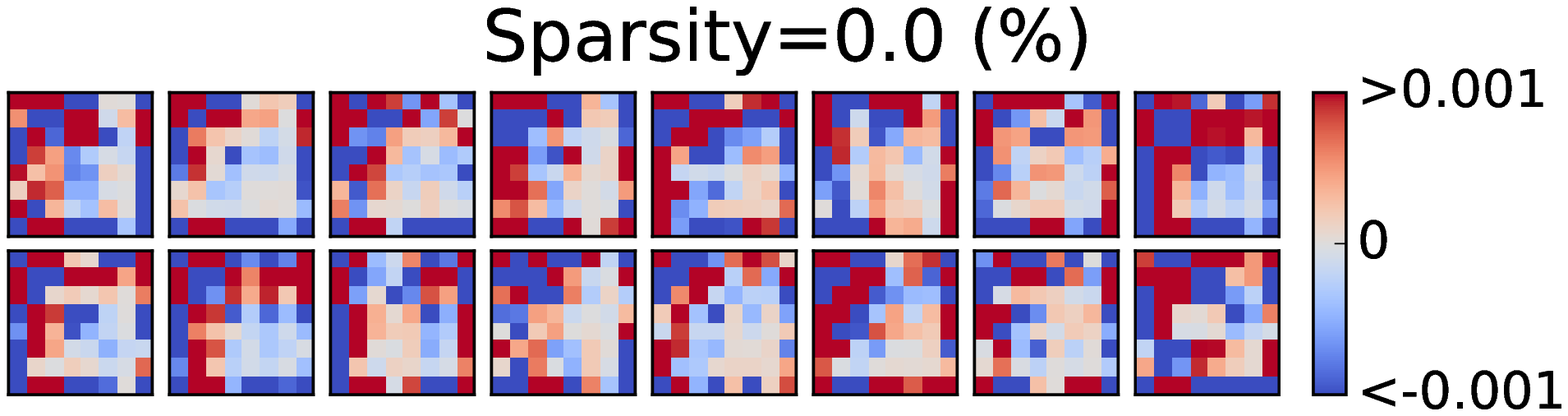} &
\includegraphics[height=0.09\textwidth]{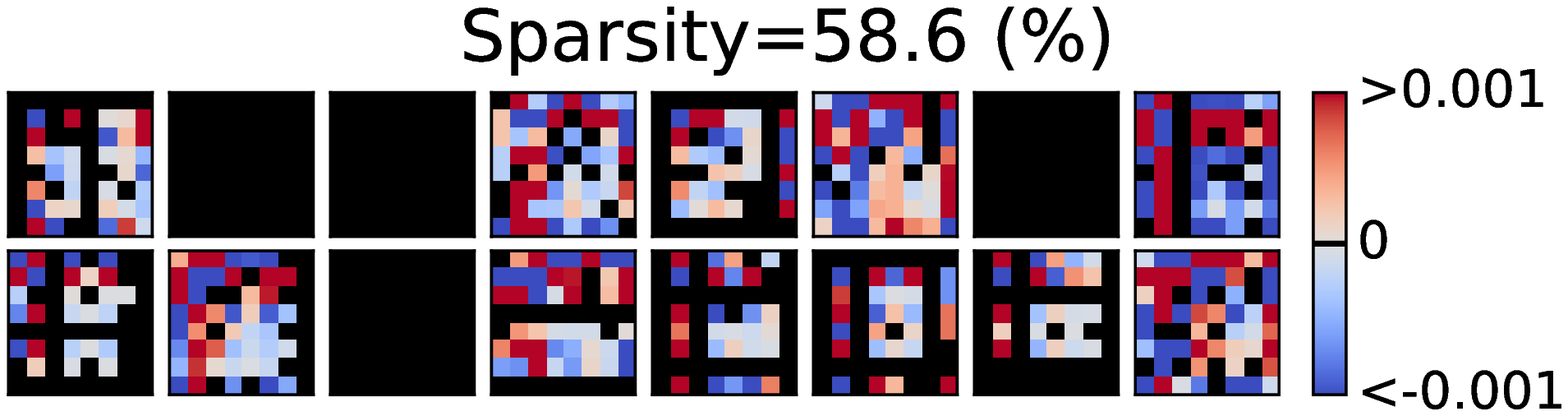} &
\includegraphics[height=0.09\textwidth]{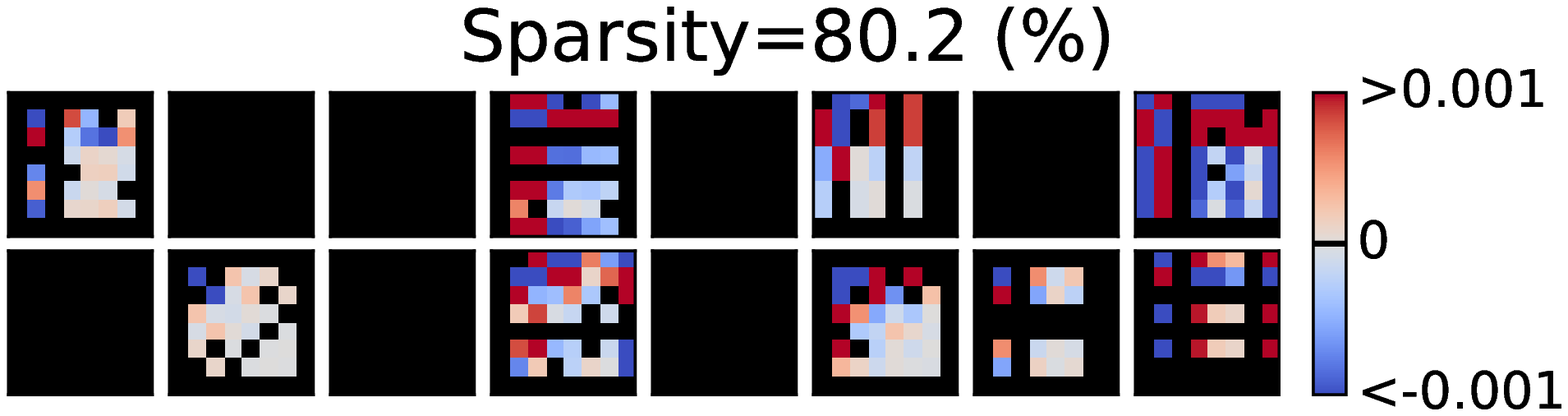} \\
\raisebox{0.6em}{\rotatebox{90}{\shortstack[c]{Spatial\\domain\\~\\~}}} &
\includegraphics[height=0.09\textwidth]{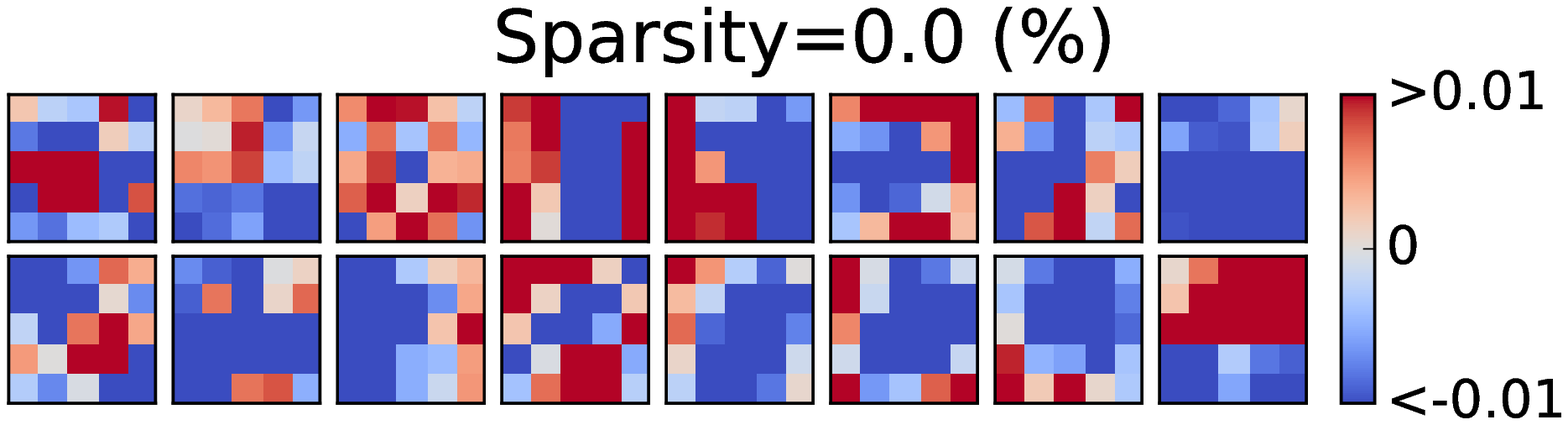} &
\includegraphics[height=0.09\textwidth]{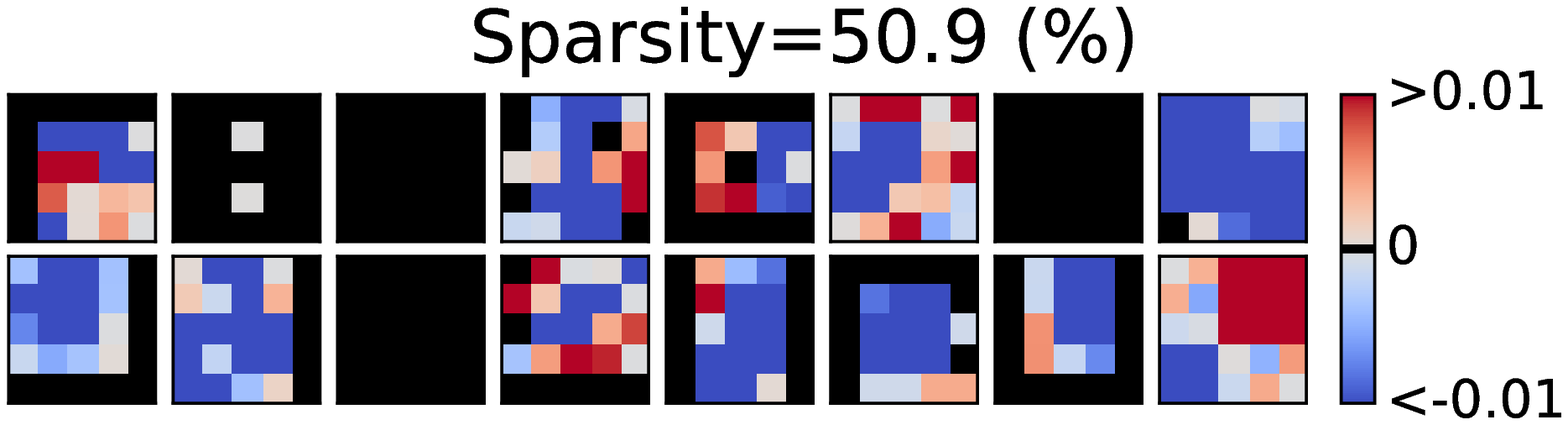} &
\includegraphics[height=0.09\textwidth]{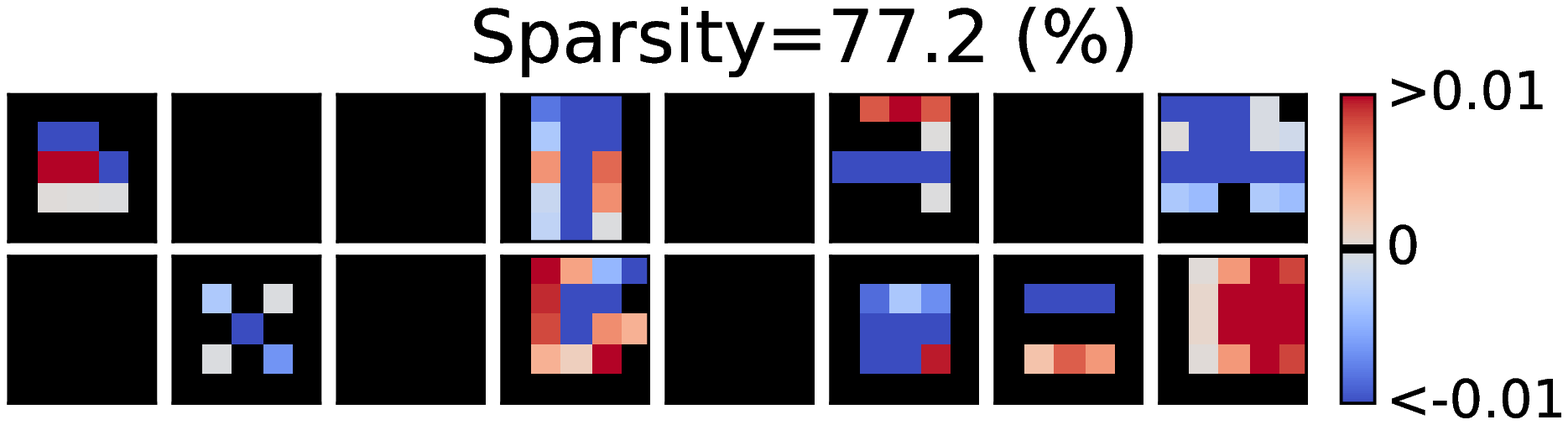} \\
\end{tabular}
}
\caption{Convolutional filter samples that are sparse after pruning either in the Winograd domain and in the spatial domain, obtained from the AlexNet second convolutional layer.\label{sec:train:fig:03}}
\end{figure}

\textbf{Examples of pruned filters}: In Figure~\ref{sec:train:fig:03}, we present convolutional filter samples that are sparse after pruning either in the Winograd domain and in the spatial domain, which are obtained by our regularization method for different sparsity levels. The AlexNet second convolutional layer consists of $5\times5$ filters and we assume to use Winograd convolution of $(r,n)=(5,8)$ in Section~\ref{sec:winograd}.

\begin{figure}[t]
\centering
\includegraphics[width=0.95\textwidth]{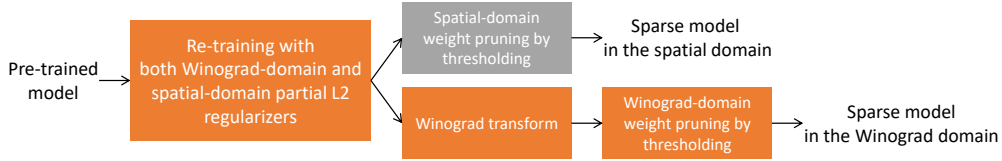}
\caption{Sparse model obtained after pruning either in the spatial domain or in the Winograd domain.\label{sec:jointsparsity:fig:01}}
\end{figure}

As observed above, we have presented our algorithms using L2 regularizers. Often L1 norms are used to promote sparsity (e.g., see \citet{chen2001atomic}), but here we suggest using L2 instead, since our goal is to induce small-value weights rather than to drive them to be really zero. The model re-trained with our L2 regularizers is still dense and not sparse before pruning. However, it is jointly regularized to have many small-value weights, which can be pruned at negligible loss, in both domains. The sparsity is actually attained only after pruning its small-value weights in either domain as shown in Figure~\ref{sec:jointsparsity:fig:01}. This is to avoid the fundamental limit of joint sparsity, similar to the uncertainty principle of the Fourier transform~\citep{donoho1989uncertainty}. See Section~\ref{sec:discussion} for more discussion.

\section{Universal compression and dual domain deployment} \label{sec:univ}

We compress the jointly sparse CNN model from Section~\ref{sec:train} by universal compression in the spatial domain for universal deployment. Universal compression consists of the following three steps, as illustrated in Figure~\ref{sec:intro:fig:01}.

\textbf{Universal quantization and pruning}: First, we randomize spatial-domain weights by adding uniform random dithers, and quantize the dithered weights uniformly with the interval of $\Delta$ by
\begin{equation} \label{sec:comp:eq:01}
q_i=\Delta\cdot\round((a_i+U_i)/\Delta),
\end{equation}
where $a_1,\dots,a_{N_{\text{SD}}}$ are the individual spatial-domain weights of all $L$ layers, and $U_1,\dots,U_{N_{\text{SD}}}$ are independent and identically distributed uniform random variables with the support of $[-\Delta/2,\Delta/2]$; the rounding function satisfies $\round(x)=\sign(x)\lfloor|x|+0.5\rfloor$, where $\lfloor x\rfloor$ is the largest integer smaller than or equal to $x$. The weights rounded to zero in \eqref{sec:comp:eq:01} are pruned and fixed to be zero for the rest of the fine-tuning and compression steps. The random dithering values or their random seed are assumed to be known at deployment, and the dithering values are cancelled for the unpruned weights after decompression by
\begin{equation} \label{sec:comp:eq:02}
\hat{q}_i
=q_i-U_i\cdot1_{q_i\neq0},
\end{equation}
where $\hat{q}_i$ is the final deployed value of weight~$a_i$ for inference. If $U_i=0$ (no dithering), this simply reduces to uniform quantization. 

\textbf{Fine-tuning the uniform codebook}: Second, we fine-tune the uniform codebook to compensate the accuracy loss after quantization. The average gradient is computed for unpruned weights that are quantized to the same value in \eqref{sec:comp:eq:01}. Then, their shared quantized value in the codebook is updated by gradient descent using the average gradient of them, which is given by
\begin{equation} \label{sec:comp:eq:03}
c_n(t)
=c_n(t-1)-\eta\frac{1}{|\mathcal{I}_n|}\sum_{i\in\mathcal{I}_n}\nabla_{a_i}C(t-1),\ \ \ n\neq0,
\end{equation}
where $t$ is the iteration time, $\eta$ is the learning rate, and $\mathcal{I}_n$ is the index set of all weights that are quantized to the same value~$c_n=n\Delta$ in \eqref{sec:comp:eq:01} for some non-zero integer~$n$. After the codebook is updated, individual weights are updated by following their shared quantized value in the codebook, i.e., 
\[
\hat{q}_i(t)=c_n(t)-U_i, \ \ \ \forall i\in\mathcal{I}_n,\ \ \ n\neq0.
\]
We emphasize here that the pruned weights in \eqref{sec:comp:eq:01} are not fine-tuned and stay zero. We do not include the spatial-domain regularizer in \eqref{sec:comp:eq:03} since this step follows after the joint sparsity optimization as shown in Figure~\ref{sec:intro:fig:01}. We determine which spatial-domain weights to prune in \eqref{sec:comp:eq:01} and fix them to zero. However, to maintain the sparsity in the Winograd-domain while optimizing the quantization codebook in the spatial domain, we keep the Winograd-domain regularizer, i.e., we use
\[
C(\mathcal{X};\mathbf{w},\zeta_{\text{WD}})
=E(\mathcal{X};\mathbf{w})
+e^{\zeta_{\text{WD}}}R_{\text{WD}}(\mathbf{w};s_{\text{WD}})
-\alpha\zeta_{\text{WD}},
\]
in \eqref{sec:comp:eq:03} instead of \eqref{sec:reg:eq:04}.

\textbf{Universal lossless source coding}: Finally, universal lossless source coding follows for compression. It is assumed that the encoder and the decoder share the information on the random dithers, or it is assumed that the dithering information can be already known to both of them through a compression protocol, e.g., by sending the random seed. The indexes in the codebook of the universally quantized weights are passed as an input stream to a universal entropy source coding scheme such as Lempel--Ziv--Welch~\citep{ziv1977universal,ziv1978compression,welch1984technique}, \textit{gzip}~\citep{gailly2003gzip} and \textit{bzip2}~\citep{seward1998bzip2} that uses the Burrows--Wheeler transform~\citep{effros2002universal}, which produces a compressed stream. We also need to deploy the codebook that contains the indexes and corresponding fine-tuned shared quantized values for decompression.

\textbf{Deployment}: At deployment, the compressed stream is decompressed, and random dithers are cancelled to get unpruned spatial-domain weights as in \eqref{sec:comp:eq:02}. Then, the CNN can be deployed in the spatial domain with the desired sparsity. If we deploy the CNN in the Winograd domain, its convolutional filters are transformed into the Winograd domain, and pruned to the desired sparsity level (see deployment in Figure~\ref{sec:intro:fig:01}).


\section{Discussion} \label{sec:discussion}

\subsection{Joint sparsity} \label{sec:jointsparsity}

The uncertainty principle for the Fourier transform establishes the fundamental limit of the joint sparsity of the signal in the time and frequency domains, e.g., see \citet[Theorem~1~and~Corollary~1]{donoho1989uncertainty} and \citet[Section 11.3.4]{eldar2015sampling}. The Winograd transform was originally proposed as a method to calculate the discrete Fourier Transform (DFT) efficiently, by reordering the input such that DFT can be implemented as cyclic convolutions~\citep{winograd1978computing}. However, reordering the input sequence by the Winograd transform does not transform it to the frequency domain. Hence, one may not directly apply the time-frequency uncertainty principles to the Winograd transform. We also show by example that some sparse spatial-domain filters can have a considerable number of zero elements even after Winograd transformation, while they become dense in the Fourier transform case, as follows:
\[
\left[\begin{array}{ccc}
0 & 0 & 0 \\
0 & 1 & 0 \\
0 & 0 & 0 \\
\end{array}\right]
\xrightarrow[\text{\shortstack[c]{Fourier\\transform}}]{}
\left[\begin{array}{ccc}
1 & e^{-2\pi i/3} & e^{-4\pi i/3} \\
e^{-2\pi i/3} & e^{-4\pi i/3} & 1 \\
e^{-4\pi i/3} & 1 & e^{-2\pi i/3} \\
\end{array}\right],
\]
and
\[
\left[\begin{array}{ccc}
0 & 0 & 0 \\
0 & 1 & 0 \\
0 & 0 & 0 \\
\end{array}\right]
\xrightarrow[\text{\shortstack[c]{Winograd\\transform}}]{}
\left[\begin{array}{cccc}
0 & 0    & 0    & 0 \\
0 &  1/4 & -1/4 & 0 \\
0 & -1/4 &  1/4 & 0 \\
0 & 0    & 0    & 0
\end{array}\right].
\]

The fundamental limit of joint sparsity explains why we use L2 instead of L1 for joint sparsity regularization. Here, we need to clarify that the model re-trained with our Winograd-domain and spatial-domain L2 regularizers is still dense and not sparse before pruning. However, it is jointly regularized to have many small-value weights, which can be pruned at negligible loss, in both domains. In other words, our regularized model is not simultaneously sparse in both domains. The sparsity is actually attained only after pruning its small-value weights in either domain, i.e., in any of the spatial or the Winograd domain (see Figure~\ref{sec:jointsparsity:fig:01}).

We further considered the compression of jointly sparse models for universal deployment. In this case, we make the model actually sparse in the spatial domain by pruning small-value weights in the spatial domain. Then, we quantize the model in the spatial domain for compression (see Figure~\ref{sec:intro:fig:01}). The resulting quantized model is sparse in the spatial domain, but it becomes dense in the Winograd domain. Thus, in order to recover the sparsity in the Winograd domain as much as possible, we fine-tune the spatial-domain quantization codebook with the Winograd-domain L2 regularizer (see Section~\ref{sec:univ}) and induce small-value Winograd-domain weights that can be pruned at small loss.

\begin{figure}[t]
\centering
{\scriptsize
\begin{tabular}{cc}
\includegraphics[height=0.25\textwidth]{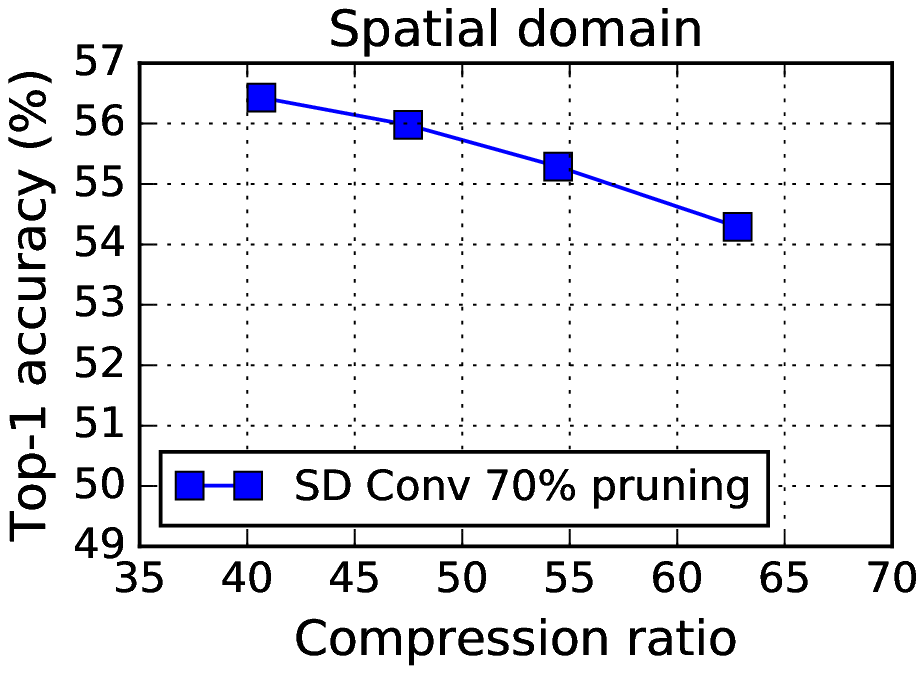} &
\includegraphics[height=0.25\textwidth]{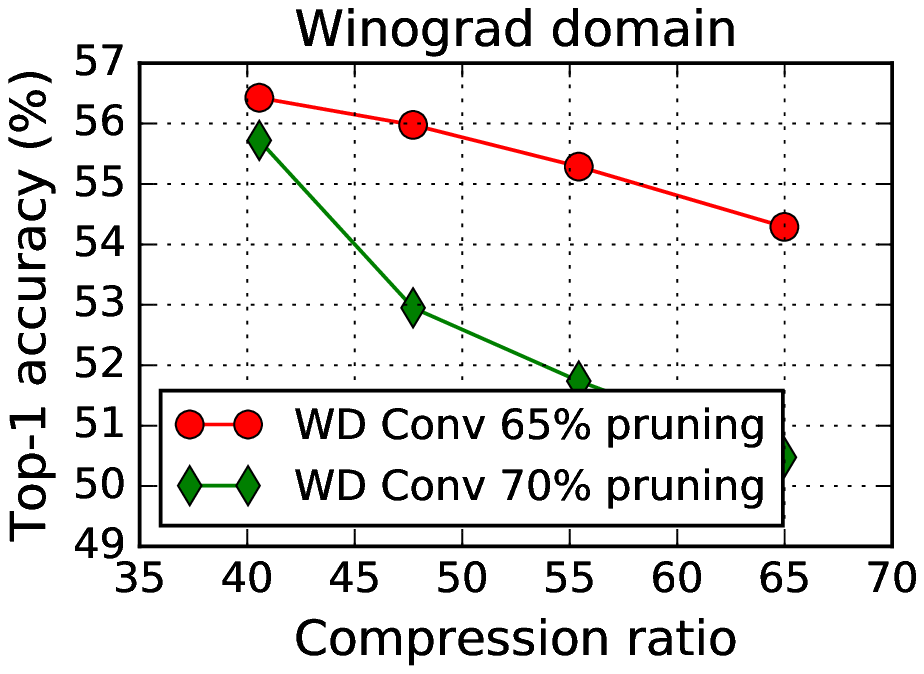} \\
\end{tabular}
}
\caption{Compression ratio versus top-1 accuracy for compressed AlexNet models on ImageNet classficiation.\label{sec:jointsparsity:fig:02}}
\end{figure}

Figure~\ref{sec:jointsparsity:fig:02} shows the compression ratio versus top-1 accuracy for compressed AlexNet models on ImageNet classification. The models are pruned, quantized, fine-tuned, and compressed, as explained above, in the spatial domain at the same pruning ratio but for different quantization cell sizes (the larger the cell size, the bigger the compression ratio). 
In the Winograd domain, we decompressed them and applied different pruning ratios to evaluate the accuracy at different sparsity levels. Observe that the accuracy degrades as the pruning ratio increases in the Winograd domain.

\begin{figure}[t]
\centering
{\scriptsize
\begin{tabular}{cc}
\includegraphics[height=0.25\textwidth]{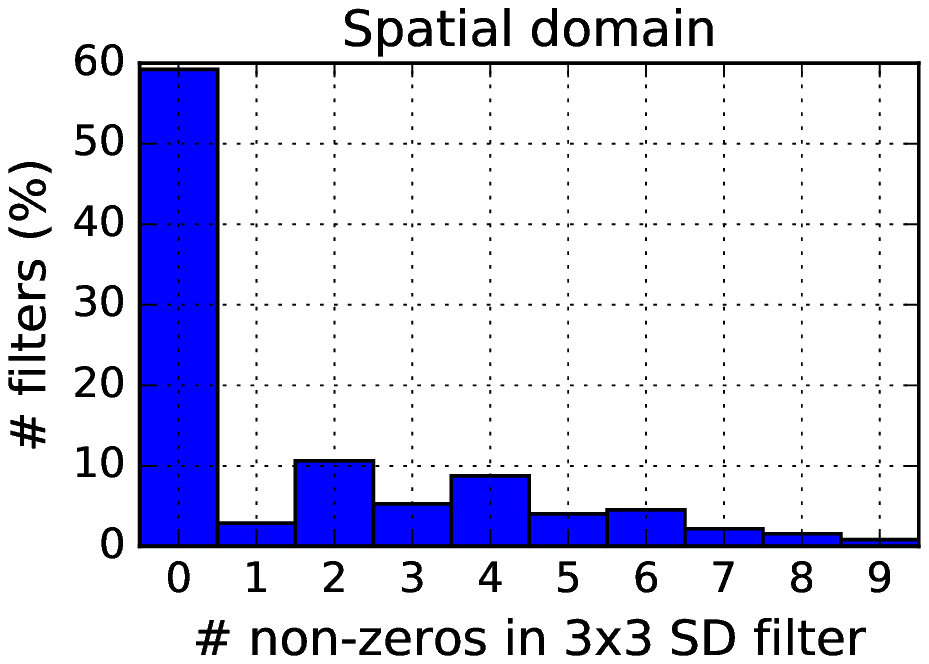} &
\includegraphics[height=0.25\textwidth]{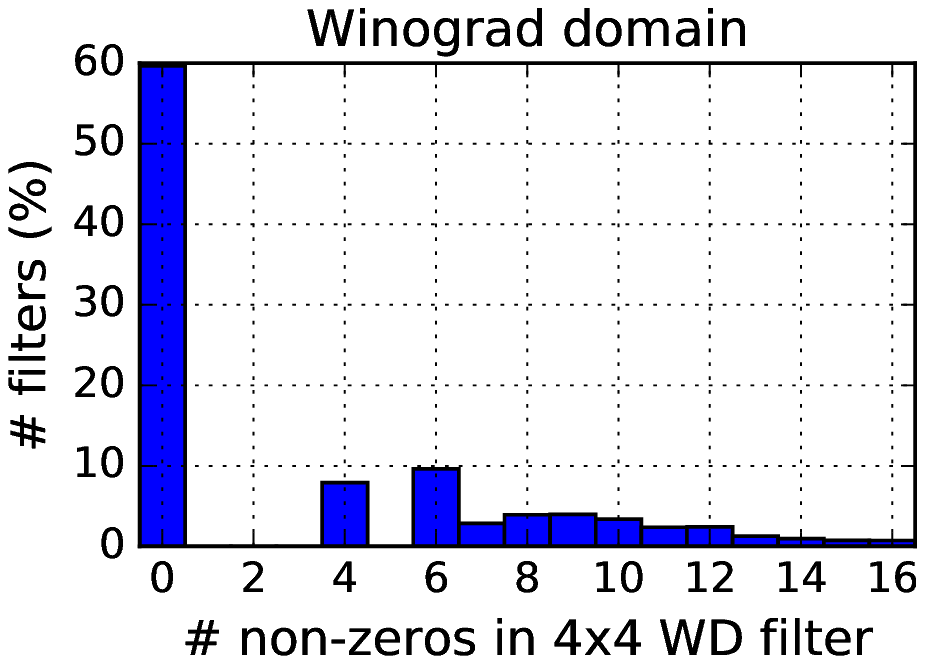} \\
\end{tabular}
}
\caption{Histogram of the number of non-zero elements in each $3\times3$ filter for our ResNet18 model targeting 80\% joint sparsity.\label{sec:jointsparsity:fig:03}}
\end{figure}

Finally, we collect the statistics of the number of non-zero elements in each $3\times3$ filter, after pruning in the spatial domain and in the Winograd domain, respectively, for our ResNet18 model targeting 80\% joint sparsity. In Figure~\ref{sec:jointsparsity:fig:03}, observe that 60\% filters are of all zeros, which can be filter-pruned, but we still have a considerable amount of sparse filters that have non-zero elements, which come from our regularization method and contribute additional 20\% sparsity.
 
\subsection{Partial L2 regularization for sparsity} \label{sec:l2}

\begin{figure}[t]
\centering
{\scriptsize
\begin{tabular}{cc}
\includegraphics[height=0.22\textwidth]{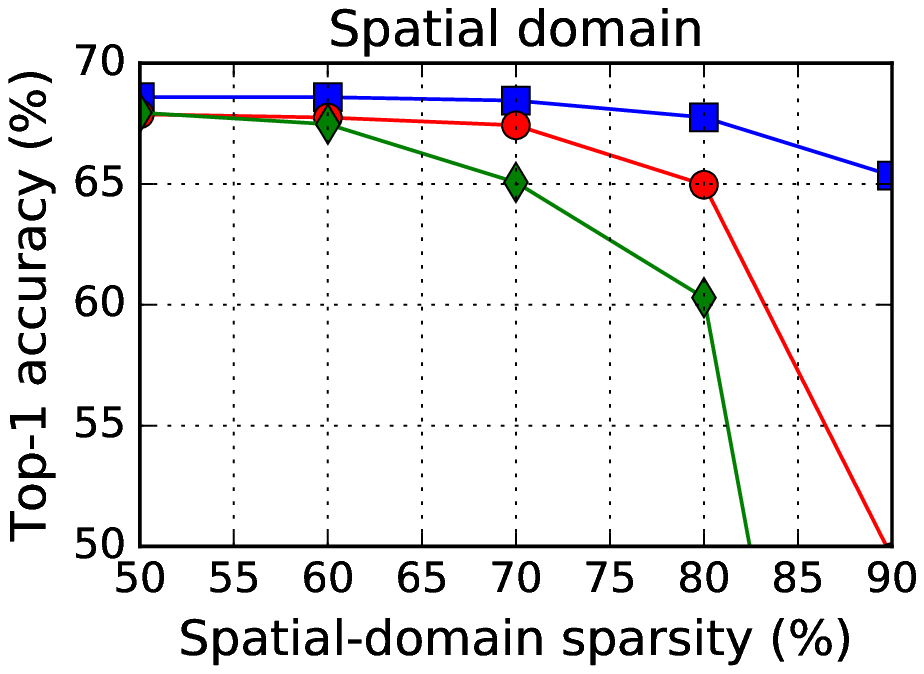} &
\includegraphics[height=0.22\textwidth]{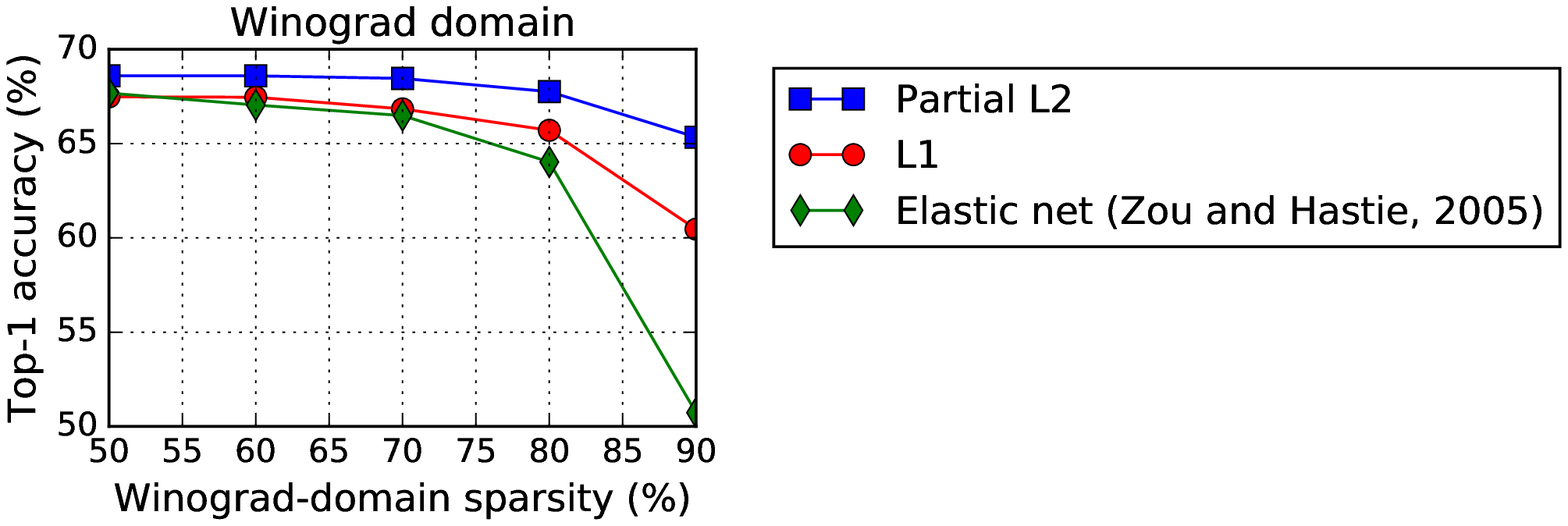} \\
\end{tabular}
}
\caption{Performance comparison of our partial L2 regularization to the conventional L1 and elastic net~\citep{zou2005regularization} regularization.\label{sec:l2:resnet18:fig:01}}
\end{figure}

We compare the performance of our partial L2 regularization method to the conventional L1 and elastic net~\citep{zou2005regularization} regularization methods for sparsity. Figure~\ref{sec:l2:resnet18:fig:01} shows the results from our experiments, and our partial L2 was better than the others empirically at least in our experiments. 
It remains as our future work to test more recent relaxation methods of sparsity constraints, such as $k$-support norm~\citep{argyriou2012sparse}.

\section{Experiments} \label{sec:exp}

\subsection{ResNet-18 for ImageNet classification} \label{sec:exp:resnet18}

\begin{figure}[t]
\centering
{\scriptsize
\begin{tabular}{cc}
\includegraphics[height=0.22\textwidth]{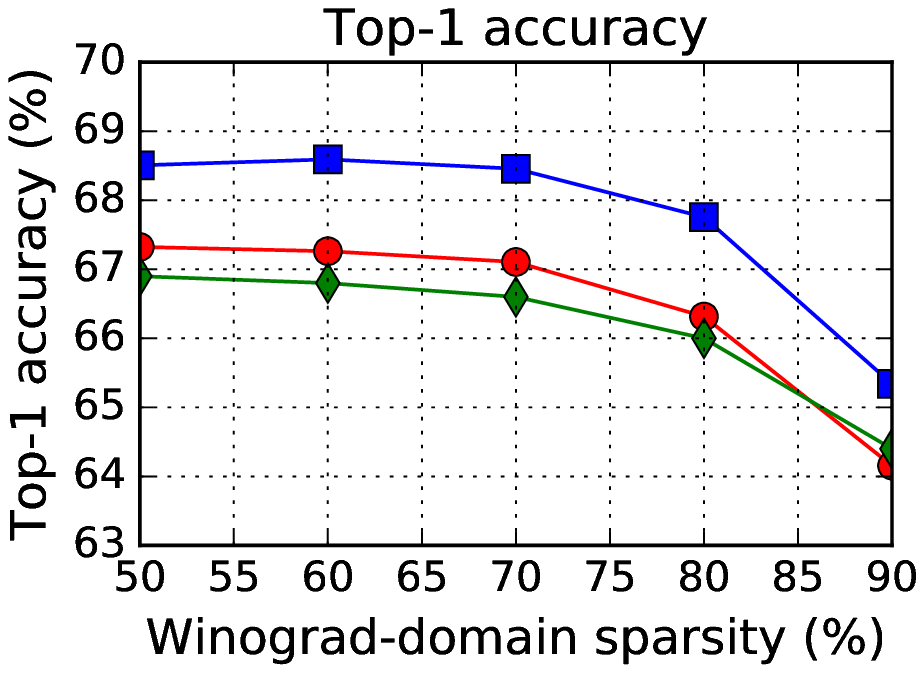} &
\includegraphics[height=0.22\textwidth]{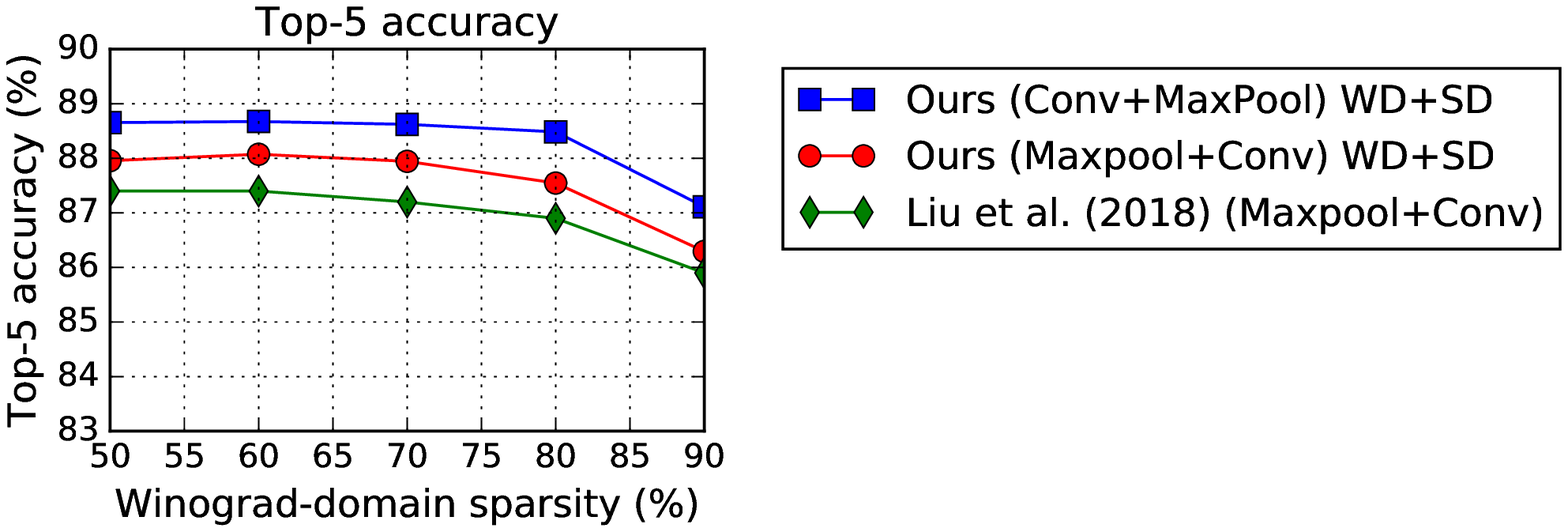} \\
\end{tabular}
}
\caption{Performance comparison of the pruned ResNet-18 models using sparse Winograd-domain convolution for different sparsity levels in the Winograd domain.\label{sec:exp:resnet18:fig:01}}
\end{figure}

We experiment our universal CNN pruning and compression scheme on the ResNet-18 model in \cite{he2016deep} for the ImageNet ILSVRC 2012 dataset~\cite{russakovsky2015imagenet}. As in \cite{liu2018efficient}, we modify the original ResNet-18 model by replacing its convolutional layers of stride $2\times2$ with convolutional layers of stride $1\times1$ and max-pooling layers, in order to utilize Winograd convolution for all possible convolutional layers. One difference from \cite{liu2018efficient} is that we place max-pooling after convolution (Conv+Maxpool) instead of placing it before convolution (Maxpool+Conv). Our modification provides better accuracy (see Figure~\ref{sec:exp:resnet18:fig:01}) although it comes with more computations.

\setlength{\tabcolsep}{0.5em}
\begin{table}[t]
\caption{Accuracy and complexity of our pruned ResNet-18 models.\label{sec:exp:resnet18:tbl:01}}
\centering
{\scriptsize
\begin{tabular}{cccccc}
\toprule
\multirow{2}{*}{\shortstack[c]{Regularization\\(sparsity~$s$)}}
 & \multirow{2}{*}{\shortstack[c]{Inference\\domain}}
 & \multirow{2}{*}{\shortstack[c]{Pruning\\ratio}}
 & \multirow{2}{*}{\shortstack[c]{Top-1 / Top-5\\accuracy}}
 & \multirow{2}{*}{\shortstack[c]{\# MACs\\per image}} \\
 \\
\midrule
Pre-trained model
               & SD & -    & 68.2 / 88.6 & 2347.1M \\
\midrule                                              
SD (80\%)      & SD & 80\% & 67.8 / 88.4 & 837.9M \\
WD (80\%)      & SD & 80\% & 44.0 / 70.5 & 819.7M  \\
WD+SD (80\%)   & SD & 80\% & 67.8 / 88.5 & 914.9M \\
\midrule
\midrule
Pre-trained model
               & WD & -    & 68.2 / 88.6 & 1174.0M \\
\midrule                                              
SD (80\%)      & WD & 80\% & 56.9 / 80.7 & 467.0M  \\
WD (80\%)      & WD & 80\% & 68.4 / 88.6 & 461.9M  \\
WD+SD (80\%)   & WD & 80\% & 67.8 / 88.5 & 522.6M  \\
\bottomrule
\end{tabular}
}
\end{table}

The Winograd-domain regularizer is applied to all $3\times3$ convolutional filters, for which Winograd convolution can be used. We assume to use Winograd convolution of $(r,n)=(3,4)$ for $3\times3$ filters (see Section~\ref{sec:winograd}). The spatial-domain regularizer is applied to all convolutional and fully-connected layers not only for pruning but also for compression later in the spatial domain. We use the Adam optimizer~\citep{kingma2014adam} with the learning rate of $1\text{e-}5$ for $500k$ iterations with the batch size of $128$. We set $\alpha=1$ in \eqref{sec:reg:eq:04}. The initial values for $\zeta_{\text{WD}}$ and $\zeta_{\text{SD}}$ are both set to be $10$, and they are updated using the Adam optimizer with the learning rate of $1\text{e-}4$.

We follow the definition of the compression ratio from \citet{han2015deep}, and it is the ratio of the original model size (without entropy coding or zipping) to the compressed model size (pruned, quantized, and entropy-coded or zipped); we used bzip2~\citep{seward1998bzip2} as our entropy coding scheme after quantizaion, instead of Huffman coding used in \citet{han2015deep}. Many of the previous DNN compression papers follow this definition, e.g., see \citet{choi2017towards,ullrich2017soft,agustsson2017soft,louizos2017bayesian,choi2018universal}, and thus we used the same one to be consistent with them. We also compared the pruning ratio and the number of MACs in our tables, which are not impacted by coding or zipping. Top-$X$ accuracy is the percentage of images whose ground-truth label is in the top-$X$ highest confidence predictions by the network (see \citet[Section~4.1]{russakovsky2015imagenet}).

Table~\ref{sec:exp:resnet18:tbl:01} summarizes the average pruning ratio, the accuracy and the number of MACs to process one input image of size $224\times224$ for pruned ResNet-18 models. The number of MACs for Winograd convolution is counted by following \citet[Section~5]{lavin2016fast}. We compare three models obtained by spatial-domain regularization only (SD), Winograd-domain regularization only (WD), and both Winograd-domain and spatial-domain regularizations (WD+SD). The accuracy is evaluated using (1) spatial-domain convolution and (2) Winograd convolution,\footnote{We used https://github.com/IntelLabs/SkimCaffe~\citep{li2017enabling} for Winograd convolution in accuracy evaluation.} for convolutional layers of $3\times3$ filters. In case of (2), the $3\times3$ filters are transformed into the Winograd domain and pruned to the desired ratio.

As expected, the proposed regularization method produces its desired sparsity only in the regularized domain. If we prune weights in the other domain, then we suffer from considerable accuracy loss. Using both Winograd-domain and spatial-domain regularizers, we can produce one model that is sparse and accurate in both domains. We can reduce the number of MACs by $2.6\times$ and $4.5\times$ when using sparse convolution in the spatial and the Winograd domains, respectively, at accuracy loss less than 0.5\%.

We compare the accuracy of our pruned ResNet-18 models to the ones from \citet{liu2018efficient} in Figure~\ref{sec:exp:resnet18:fig:01}. Observe that our models outperform the ones from \citet{liu2018efficient} in the Winograd domain. We emphasize that the major advantage of our scheme is that it produces one model that can use any of sparse spatial-domain convolution or sparse Winograd convolution. However, the models from \citet{liu2018efficient} are constrained to utilize their special Winograd-ReLU layers even though they can additionally exploit the dynamic sparsity of ReLU activations in the Winograd domain as explained in Section~\ref{sec:sparse}.

\begin{figure}
\centering
\includegraphics[width=0.92\textwidth]{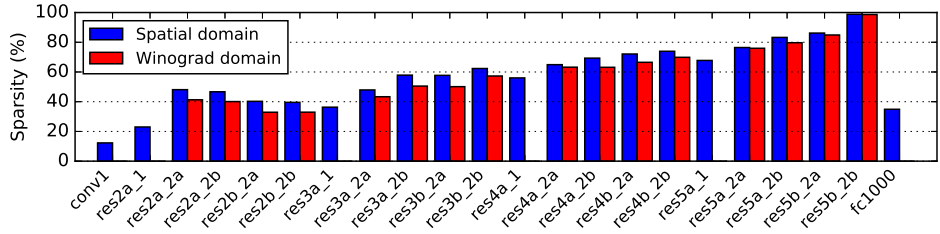}
\caption{Sparsity by layer of our jointly sparse ResNet-18 model, i.e., WD+SD (80\%) in Table~1.\label{sec:exp:resnet18:fig:02}}
\end{figure}

In Figure~\ref{sec:exp:resnet18:fig:02}, we present the layer-by-layer sparsity of our jointly sparse ResNet-18 model, obtained by the regularization of WD+SD (80\%) in Table~1 of our paper. Observe that the Winograd-domain sparsity is provided for convolutional layers of $3\times3$ filters only, where Winograd convolution can be used, while the spatial-domain sparsity is given for all layers. We have different pruning ratios by layer since we use one threshold value for pruning all layers (see Section~3.1), in contrast to \citet{liu2018efficient}, where all convolutional layers of $3\times3$ filters are pruned at the same ratio in the Winograd domain. Allowing different pruning ratios by layer can be beneficial since some layers can be more important than the others and we may want to prune less in the important layers.

\setlength{\tabcolsep}{0.5em}
\begin{table}[t]
\caption{Summary of compression results for the ResNet-18 model.\label{sec:exp:resnet18:tbl:02}}
\centering
{\scriptsize
\begin{tabular}{cccccc}
\toprule
\multirow{2}{*}{\shortstack[c]{Regularization\\(sparsity~$s$)}}
 & \multirow{2}{*}{\shortstack[c]{Quantization\\(cell size~$\Delta$)}}
 & \multirow{2}{*}{\shortstack[c]{Compression\\ratio}}
 & \multirow{2}{*}{\shortstack[c]{Inference\\domain}}
 & \multirow{2}{*}{\shortstack[c]{Top-1 / Top-5\\accuracy}}
 & \multirow{2}{*}{\shortstack[c]{\# MACs\\per image}} \\
 \\
\midrule
\multicolumn{2}{c}{Pre-trained model}
               & -    & SD   & 68.2 / 88.6 & 2347.1M \\
\midrule
\multirow{6}{*}{\shortstack[c]{WD+SD\\(80\%)}}
 & UQ (0.005)  & 24.2 & SD   & 67.4 / 88.2 & 888.6M \\
 & UQ (0.010)  & 28.9 & SD   & 66.9 / 87.9 & 859.0M \\
 & UQ (0.020)  & 38.4 & SD   & 63.7 / 86.0 & 749.6M \\
\cmidrule{2-6}               
 & DUQ (0.005) & 23.8 & SD   & 67.5 / 88.2 & 886.5M \\
 & DUQ (0.010) & 28.7 & SD   & 66.8 / 87.8 & 848.1M \\
 & DUQ (0.020) & 38.6 & SD   & 60.0 / 83.5 & 708.8M \\
\midrule
\midrule
\multicolumn{2}{c}{Pre-trained model}
               & -    & WD   & 68.2 / 88.6 & 1174.0M \\
\midrule
\multirow{6}{*}{\shortstack[c]{WD+SD\\(80\%)}}
 & UQ (0.005)  & 24.2 & WD   & 67.4 / 88.2 & 516.4M \\
 & UQ (0.010)  & 28.9 & WD   & 66.9 / 87.9 & 516.5M \\
 & UQ (0.020)  & 38.4 & WD   & 63.7 / 86.0 & 495.1M \\
\cmidrule{2-6}               
 & DUQ (0.005) & 23.8 & WD   & 67.4 / 88.3 & 516.3M \\
 & DUQ (0.010) & 28.7 & WD   & 66.6 / 87.7 & 512.9M \\
 & DUQ (0.020) & 38.6 & WD   & 60.0 / 83.5 & 502.5M \\
\bottomrule
\end{tabular}
}
\end{table}

Table~\ref{sec:exp:resnet18:tbl:02} shows our universal CNN quantization and compression results for the ResNet-18 model. We take the model obtained by regularization of WD+SD (80\%) in Table~\ref{sec:exp:resnet18:tbl:01} and compress its weights as described in Section~\ref{sec:univ}. We compare uniform quantization (UQ) and dithered uniform quantization (DUQ). We use \textit{bzip2}~\citep{seward1998bzip2} for universal source coding. The results show that we can achieve more than $24\times$ compression at accuracy loss less than 1\% in both cases (1) and (2). 

\subsection{AlexNet for ImageNet classification} \label{sec:exp:alexnet}

We perform similar pruning and compression experiments for AlexNet~\citep{krizhevsky2012imagenet}. The AlexNet model has a huge number of weights in its fully-connected (FC) layers (58.6M out of total 61M), and thus we first prune roughly 90\% spatial-domain weights mostly from its FC layers by the incremental pruning as suggested in \citet{han2015learning}. 

We re-train the pruned AlexNet model, similar to the ResNet-18 case above. In particular, we apply the proposed regularizers only to the second to the fifth convolutional layers (Conv2-Conv5), where their filter sizes are small such as $3\times3$ and $5\times5$. We assume to use Winograd convolution of $(r,n)=(3,6)$ and $(r,n)=(5,8)$ for $3\times3$ filters and $5\times5$ filters, respectively. 
The first convolutional layer (Conv1) is excluded since its filter size is $11\times11$, which is not small for Winograd convolution.

\setlength{\tabcolsep}{0.5em}
\begin{table}[t]
\caption{Summary of compression results for the AlexNet model.\label{sec:exp:alexnet:tbl:01}}
\centering
{\scriptsize
\begin{tabular}{cccccc}
\toprule
\multirow{2}{*}{\shortstack[c]{Regularization\\(sparsity~$s$)}}
 & \multirow{2}{*}{\shortstack[c]{Quantization\\(cell size~$\Delta$)}}
 & \multirow{2}{*}{\shortstack[c]{Compression\\ratio}}
 & \multirow{2}{*}{\shortstack[c]{Inference\\domain*}}
 & \multirow{2}{*}{\shortstack[c]{Top-1 / Top-5\\accuracy (\%)}}
 & \multirow{2}{*}{\shortstack[c]{\# MACs\\per image}} \\ \\
\midrule
\multicolumn{2}{c}{Pre-trained model}
               & -    & SD   & 56.8 / 80.0 & 724.4M \\
\midrule
\multirow{6}{*}{\shortstack[c]{WD+SD\\(70\%)}}
 & UQ (0.005)  & 40.7 & SD   & 56.4 / 79.7 & 253.7M \\
 & UQ (0.010)  & 47.5 & SD   & 56.0 / 79.5 & 237.1M \\
 & UQ (0.020)  & 62.8 & SD   & 54.3 / 78.0 & 211.3M \\
\cmidrule{2-6}
 & DUQ (0.005) & 40.7 & SD   & 56.4 / 79.7 & 256.1M \\
 & DUQ (0.010) & 47.7 & SD   & 56.1 / 79.3 & 240.0M \\
 & DUQ (0.020) & 65.0 & SD   & 52.8 / 77.1 & 213.5M \\
\midrule
\multicolumn{2}{c}{\citet{han2015deep}}
               & 35.0 & SD   & 57.2 / 80.3 & 301.1M \\
\multicolumn{2}{c}{\citet{guo2016dynamic}}
               & N/A  & SD   & 56.9 / 80.0 & 254.2M \\
\midrule
\midrule
\multicolumn{2}{c}{Pre-trained model}
               & -    & WD   & 56.8 / 80.0 & 330.0M \\
\midrule
\multirow{6}{*}{\shortstack[c]{WD+SD\\(70\%)}}
 & UQ (0.005)  & 40.7 & WD   & 56.4 / 79.7 & 146.2M \\
 & UQ (0.010)  & 47.5 & WD   & 56.0 / 79.5 & 144.2M \\
 & UQ (0.020)  & 62.8 & WD   & 54.3 / 78.0 & 134.9M \\
\cmidrule{2-6}
 & DUQ (0.005) & 40.7 & WD   & 56.4 / 79.7 & 145.7M \\
 & DUQ (0.010) & 47.7 & WD   & 56.0 / 79.3 & 142.6M \\
 & DUQ (0.020) & 65.0 & WD   & 52.8 / 77.0 & 132.6M \\
\midrule
\multicolumn{2}{c}{\citet{li2017enabling}}
               & N/A  & WD   & 57.3 / N/A  & 319.8M \\
\bottomrule
\multicolumn{6}{r}{* Winograd convolution is used for Conv2--Conv5 in WD inference.}
\end{tabular}
}
\end{table}

\setlength{\tabcolsep}{0.5em}
\begin{table}[t]
\caption{Summary of layer-by-layer sparsity for the AlexNet models in Table~\ref{sec:exp:alexnet:tbl:01}.\label{sec:exp:alexnet:tbl:02}}
\centering
{\scriptsize
\begin{tabular}{ccccccccccc}
\toprule
\multirow{2}{*}{\shortstack[c]{Regularization\\(sparsity~$s$)}}
 & \multirow{2}{*}{\shortstack[c]{Quantization\\(cell size~$\Delta$)}}
 & \multirow{2}{*}{\shortstack[c]{Inference\\domain*}}
 & \multicolumn{8}{c}{Sparsity (\%)} \\
\cmidrule{4-11}
 & & & Conv1 & Conv2 & Conv3 & Conv4 & Conv5 & FC1 & FC2 & FC3 \\
\midrule
\multirow{6}{*}{\shortstack[c]{WD+SD\\(70\%)}}
 & UQ (0.005)  & SD   & 15.7 & 62.9 & 81.2 & 75.2 & 71.9 & 93.2 & 92.1 & 80.2 \\
 & UQ (0.010)  & SD   & 17.2 & 68.3 & 81.9 & 76.1 & 72.7 & 93.5 & 92.2 & 80.4 \\
 & UQ (0.020)  & SD   & 25.4 & 73.8 & 83.0 & 77.5 & 73.9 & 94.9 & 93.1 & 81.7 \\
\cmidrule{2-11}
 & DUQ (0.005) & SD   & 15.8 & 62.2 & 80.9 & 74.8 & 71.6 & 93.2 & 92.1 & 80.2 \\
 & DUQ (0.010) & SD   & 18.3 & 66.8 & 81.7 & 75.8 & 72.4 & 93.7 & 92.3 & 80.6 \\
 & DUQ (0.020) & SD   & 26.8 & 71.9 & 83.1 & 77.5 & 73.9 & 95.4 & 93.6 & 82.6 \\
\midrule
\multicolumn{2}{c}{\citet{han2015deep}}
               & SD   & 16.0 & 62.0 & 65.0 & 63.0 & 63.0 & 91.0 & 91.0 & 75.0 \\
\multicolumn{2}{c}{\citet{guo2016dynamic}}
               & SD   & 46.2 & 59.4 & 71.0 & 67.7 & 67.5 & 96.3 & 93.4 & 95.4 \\
\midrule
\midrule
\multirow{6}{*}{\shortstack[c]{WD+SD\\(70\%)}}
 & UQ (0.005)  & WD   & 15.7 & 43.6 & 72.0 & 63.7 & 62.5 & 93.2 & 92.1 & 80.2 \\
 & UQ (0.010)  & WD   & 17.2 & 43.9 & 72.0 & 63.7 & 62.4 & 93.5 & 92.2 & 80.4 \\
 & UQ (0.020)  & WD   & 25.4 & 45.2 & 72.0 & 63.6 & 62.1 & 94.9 & 93.1 & 81.7 \\
\cmidrule{2-11}
 & DUQ (0.005) & WD   & 15.8 & 47.4 & 71.7 & 63.3 & 62.0 & 93.2 & 92.1 & 80.2 \\
 & DUQ (0.010) & WD   & 18.3 & 47.4 & 71.7 & 63.3 & 62.0 & 93.7 & 92.3 & 80.6 \\
 & DUQ (0.020) & WD   & 26.8 & 45.7 & 71.9 & 63.5 & 62.0 & 95.4 & 93.6 & 82.6 \\
\midrule
\multicolumn{2}{c}{\citet{li2017enabling}}
               & WD   & 0.0  & 90.6 & 95.8 & 94.3 & 93.9 & 0.0  & 0.0  & 0.0 \\
\bottomrule
\multicolumn{11}{r}{* Winograd convolution is used for Conv2--Conv5 in WD inference.}
\end{tabular}
}
\end{table}

In Table~\ref{sec:exp:alexnet:tbl:01} and Table~\ref{sec:exp:alexnet:tbl:02}, we provide the compression ratio, the sparsity, the accuracy and the number of MACs to process one input image of size $227\times227$ for compressed AlexNet models. We compare our results to \citet{han2015deep,guo2016dynamic} in the spatial domain and to \citet{li2017enabling} in the Winograd domain. We note that \citet{han2015deep,guo2016dynamic,li2017enabling} produce sparse models only in one domain. The AlexNet model in \citet{guo2016dynamic} has more pruning in FC layers and less pruning in Conv layers than ours. Hence, the overall pruning ratio is larger in \citet{guo2016dynamic} (since FC layers are dominant in the number of parameters), but ours results in more computational cost reduction (since Conv layers are dominant in the number of MACs). Furthermore, our model can be even sparse in the Winograd domain.
Comparing to \citet{li2017enabling}, our method yields less pruning for the Conv2-Conv5 layers in the Winograd domain, but we also prune the Conv1 and FC layers heavily in the spatial domain. Observe that we can reduce the number of MACs by $3.0\times$ and $5.1\times$ when using sparse spatial-domain convolution and using sparse Winograd convolution, respectively, at accuracy loss less than 1\%. The results also show that we can achieve more than $47\times$ compression.

\subsection{CT-SRCNN for image super resolution} \label{sec:exp:ct-srcnn9}

\setlength{\tabcolsep}{0.5em}
\begin{table}[t]
\caption{Summary of compression results for the 9-layer CT-SRCNN model of up-scaling factor~$3$.\label{sec:exp:ctsrcnn9:tbl:01}}
\centering
{\scriptsize
\begin{tabular}{ccccccc}
\toprule
\multirow{2}{*}{\shortstack[c]{Regularization\\(sparsity~$s$)}}
 & \multirow{2}{*}{\shortstack[c]{Quantization\\(cell size~$\Delta$)}}
 & \multirow{2}{*}{\shortstack[c]{Compression\\ratio}}
 & \multirow{2}{*}{\shortstack[c]{Inference\\domain}}
 & PSNR (dB)
 & SSIM
 & \multirow{2}{*}{\shortstack[c]{\# MACs\\per image}} \\
 \\
\midrule
\multicolumn{2}{c}{Pre-trained model}
               & -    & SD   & 29.70 & 0.8301 & 233.2G \\
\midrule
\multirow{6}{*}{\shortstack[c]{WD+SD (90\%)}}
 & UQ (0.005)  & 27.2 & SD   & 29.39 & 0.8236 & 21.1G \\
 & UQ (0.010)  & 30.5 & SD   & 29.38 & 0.8237 & 19.7G \\
 & UQ (0.020)  & 35.4 & SD   & 29.32 & 0.8225 & 17.4G \\
\cmidrule{2-7}
 & DUQ (0.005) & 27.1 & SD   & 29.38 & 0.8234 & 21.1G \\
 & DUQ (0.010) & 30.3 & SD   & 29.37 & 0.8233 & 19.8G \\
 & DUQ (0.020) & 34.8 & SD   & 29.30 & 0.8222 & 18.0G \\
\midrule
\midrule
\multicolumn{2}{c}{Pre-trained model}
               & -    & WD   & 29.70 & 0.8301 & 56.7G \\
\midrule
\multirow{6}{*}{\shortstack[c]{WD+SD (90\%)}}
 & UQ (0.005)  & 27.2 & WD   & 29.38 & 0.8235 & 10.7G \\
 & UQ (0.010)  & 30.5 & WD   & 29.38 & 0.8237 & 10.3G \\
 & UQ (0.020)  & 35.4 & WD   & 29.32 & 0.8225 & 9.9G  \\
\cmidrule{2-7}
 & DUQ (0.005) & 27.1 & WD   & 29.37 & 0.8232 & 10.7G \\
 & DUQ (0.010) & 30.3 & WD   & 29.37 & 0.8233 & 10.3G \\
 & DUQ (0.020) & 34.8 & WD   & 29.31 & 0.8222 & 10.0G \\
\bottomrule
\end{tabular}
}
\end{table}

We finally evaluate the proposed scheme for the cascade-trained SRCNN (CT-SRCNN) model of 9 convolutional layers~\citep{ren2018ctsrcnn}. We apply the Winograd-domain regularizer to the $3\times3$ and $5\times5$ filters of the second to the last layers; the $9\times9$ filters of the first layer are excluded. We use Winograd convolution of $(r,n)=(3,6)$ and $(r,n)=(5,8)$ for $3\times3$ and $5\times5$ filters, respectively. The spatial-domain regularizer is applied to all 9 layers. 

The average peak-signal-to-noise-ratio (PSNR) and structured-similarity (SSIM) are compared for Set14 dataset~\citep{zeyde2010single} in Table~\ref{sec:exp:ctsrcnn9:tbl:01} for compressed CT-SRCNN models. We also summarize the compression ratio and the number of MACs for super resolution to get one high-definition image of size $1920\times1080$ by up-scaling factor~$3$ in Table~\ref{sec:exp:ctsrcnn9:tbl:01}. Observe that we achieve $35\times$ compression at PSNR loss less than $0.4$ dB. The number of MACs is reduced by $13.4\times$ and $23.6\times$ when using sparse spatial-domain convolution and using sparse Winograd convolution, respectively.

\section{Conclusion} \label{sec:conclusion}

We introduced a framework for hardware or software platform independent pruning and compression of CNNs. The proposed scheme produces one compressed model whose convolutional filters can be made sparse either in the Winograd domain or in the spatial domain with minimal loss of accuracy and without further training. Thus, one compressed model can be deployed on any platform and the sparsity of its convolutional filters can be utilized for complexity reduction in either domain, unlike the previous approaches that yield sparse models in one domain only. We showed by experiments that the proposed scheme successfully prunes and compresses ResNet-18, AlexNet and 9-layer CT-SRCNN with compression ratios of $24.2\times$, $47.7\times$ and $35.4\times$, while reducing complexity by $4.5\times$, $5.1\times$ and $23.5\times$, respectively. Finally, our regularization method for joint sparsity can be extended for sparse frequency-domain convolution, which remains as our future work. It will be also interesting to compare our partial L2 norm to $k$-support norm~\citep{argyriou2012sparse} for sparsity regularization.




\newpage

\appendix
\section*{Appendix} \label{sec:app1}

In this appendix, we show the proof of \eqref{sec:reg:eq:06} of our paper.
\begin{proof}
We have (e.g., see \cite[Section~1.3.7]{golub2012matrix})
\begin{equation}
\label{sec:app1:eq:01}
vec(AXB)=(B^T\otimes A)vec(X),
\end{equation}
where $vec(X)$ is the column-vectorization of matrix~$X$ and $\otimes$ denotes the Kronecker product of two matrices. Thus, it follows that
\begin{equation}
\label{sec:app1:eq:02}
\|(GwG^T)\odot1_{|GwG^T|\leq\theta}\|^2
=\|((G\otimes G)vec(w))\odot1_{|(G\otimes G)vec(w)|\leq\theta}\|^2.
\end{equation}
For any matrix~$A$, column vector~$x$ and column vector~$y$, it is straightforward to show that
\[
\|(Ax)\odot y\|^2
=\|\text{diag}(y)Ax\|^2
=x^TA^T\text{diag}^2(y)Ax,
\]
where $\text{diag}(y)$ is the diagonal matrix whose diagonal elements are from $y$, and then it follows that
\begin{equation}
\label{sec:app1:eq:03}
\nabla_{x}\|(Ax)\odot y\|^2
=\nabla_{x}(x^TA^T\text{diag}^2(y)Ax)
=2A^T\text{diag}^2(y)Ax
=2A^T(Ax\odot y\odot y).
\end{equation}
Combining \eqref{sec:app1:eq:01}--\eqref{sec:app1:eq:03} leads us to
\begin{equation}
\label{sec:app1:eq:04}
\nabla_{w}\|(GwG^T)\odot1_{|GwG^T|\leq\theta}\|^2
=2G^T((GwG^T)\odot1_{|GwG^T|\leq\theta})G,
\end{equation}
which results in \eqref{sec:reg:eq:06} of our paper.
\end{proof}
We note that the gradient is actually not defined at the discontinuous points where any element in $GwG^T$ is exactly equal to the threshold~$\theta$ in magnitude, which however can be ignored in stochastic gradient descent.

{\small
\newcommand{\BIBdecl}{\setlength{\itemsep}{0.25em}}
\bibliographystyle{unsrtnat}
\bibliography{ref}
}

\end{document}